\documentclass[10pt,twocolumn,letterpaper]{article}

\usepackage{wacv}
\usepackage{times}
\usepackage{epsfig}
\usepackage{graphicx}
\usepackage{amsmath}
\usepackage{amssymb}
\usepackage{booktabs}
\usepackage{multirow}
\DeclareUnicodeCharacter{FFFC}{ }

\usepackage{tabularx}
\usepackage{cuted}
\usepackage{capt-of}
\usepackage{array}
\newcolumntype{?}{!{\vrule width 0.8pt}}
\usepackage [english]{babel}
\usepackage [autostyle, english = american]{csquotes}
\MakeOuterQuote{"}

\def\wacvPaperID{452} % Enter the WACV Paper ID here

\wacvfinalcopy % *** Uncomment this line for the final submission

\ifwacvfinal
\def\assignedStartPage{9876} % *** Enter the assigned starting page number (instead of 9876)
\fi

\ifwacvfinal
\usepackage[breaklinks=true,bookmarks=false]{hyperref}
\else
\usepackage[pagebackref=true,breaklinks=true,colorlinks,bookmarks=false]{hyperref}
\fi

\ifwacvfinal
\else
\fi

\begin{document}

%%%%%%%%% TITLE
\title{End-to-End Chinese Landscape Painting Creation Using Generative Adversarial Networks}

\author{Alice Xue\\
Princeton University\\
Princeton, NJ 08544\\
{\tt\small axue@princeton.edu}
}

\maketitle

\begin{strip}\centering
\setlength\tabcolsep{3pt}%%
\begin{tabular}{cccc}
 \shortstack{(a) Human} &
 \shortstack{(b) Baselines} &
 \shortstack{(c) Ours\\(RaLSGAN+Pix2Pix)} &
 \shortstack{(d) Ours\\(StyleGAN2+Pix2Pix)} \\
 \includegraphics[width=0.2\textwidth]{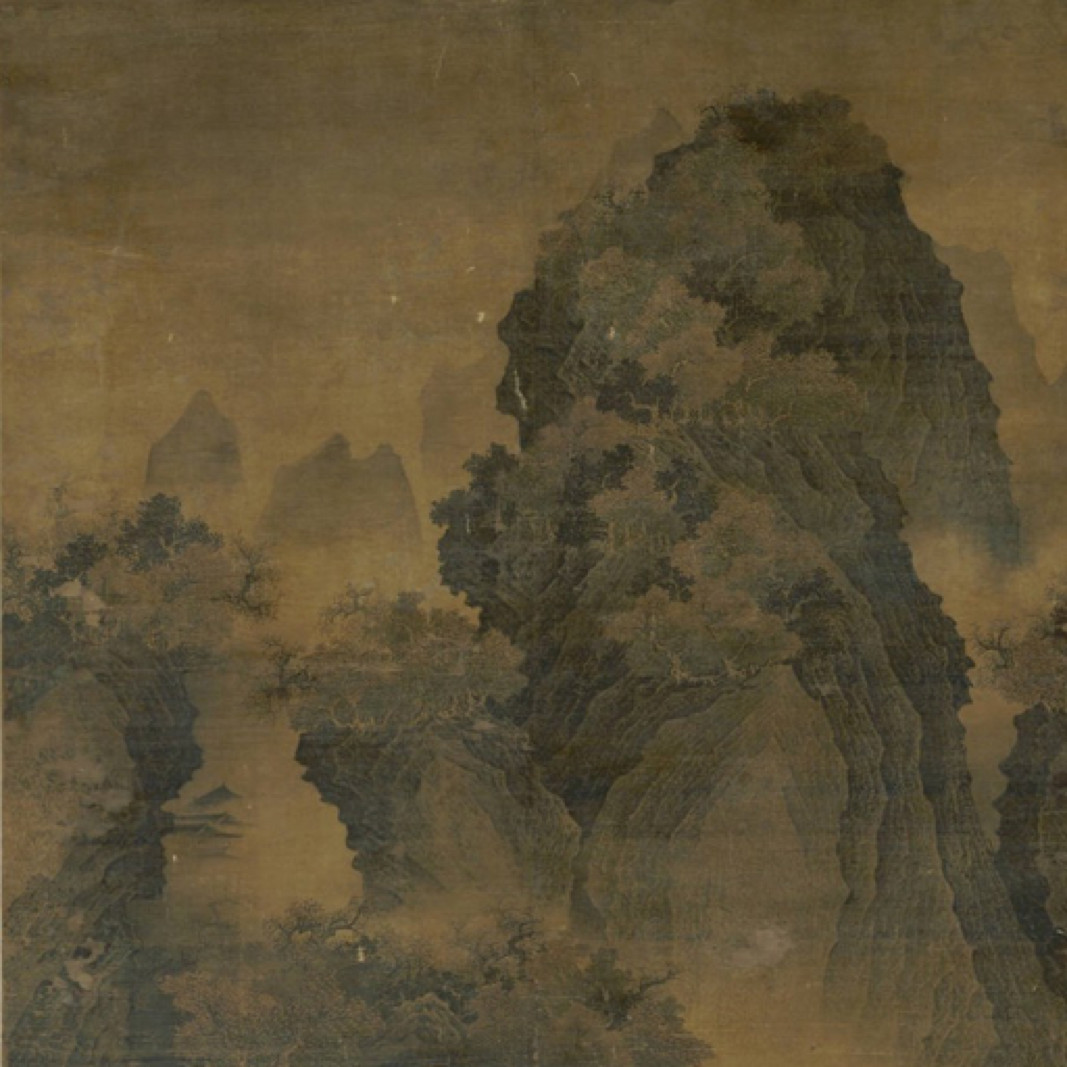} &
 \includegraphics[width=0.2\textwidth]{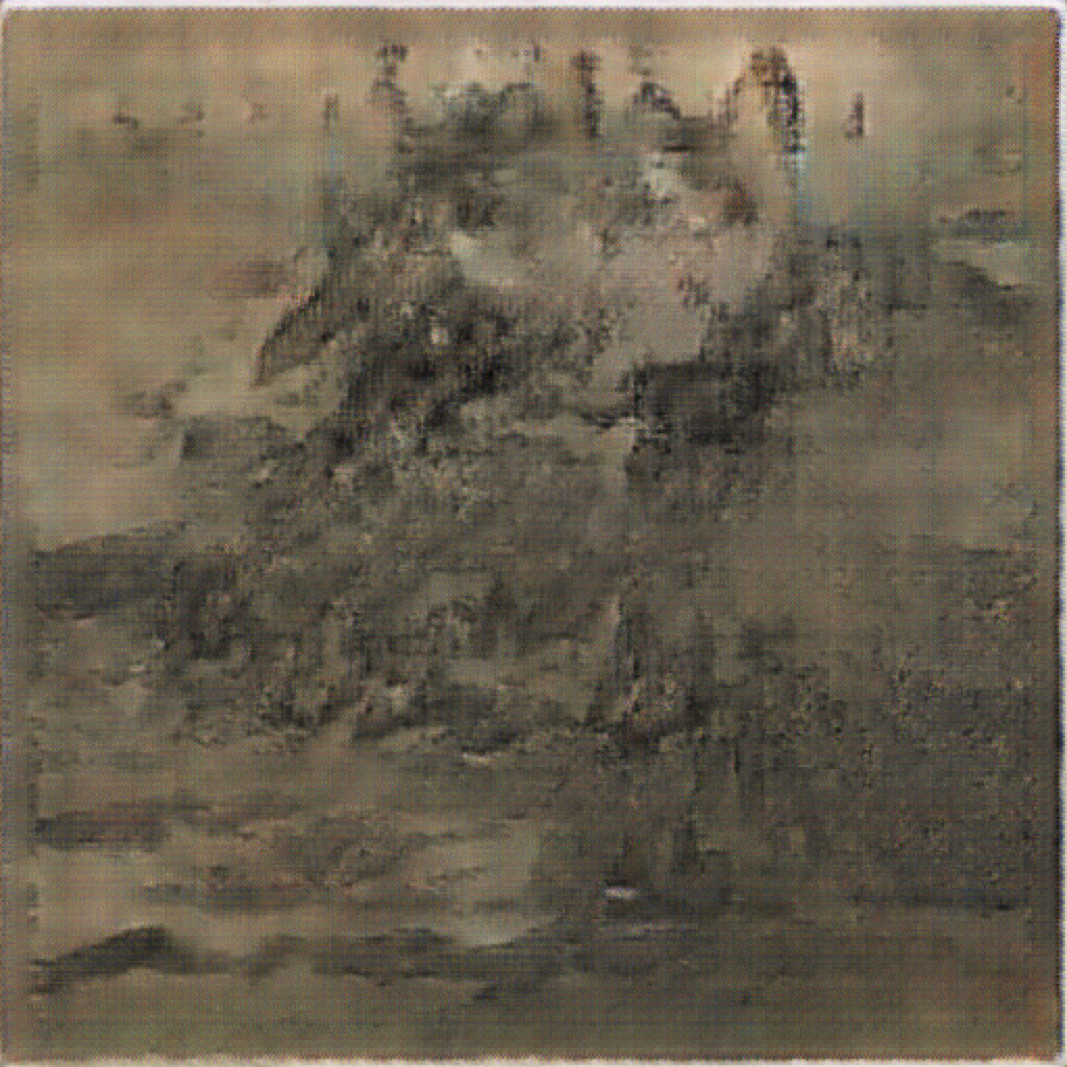} &
 \includegraphics[width=0.2\textwidth]{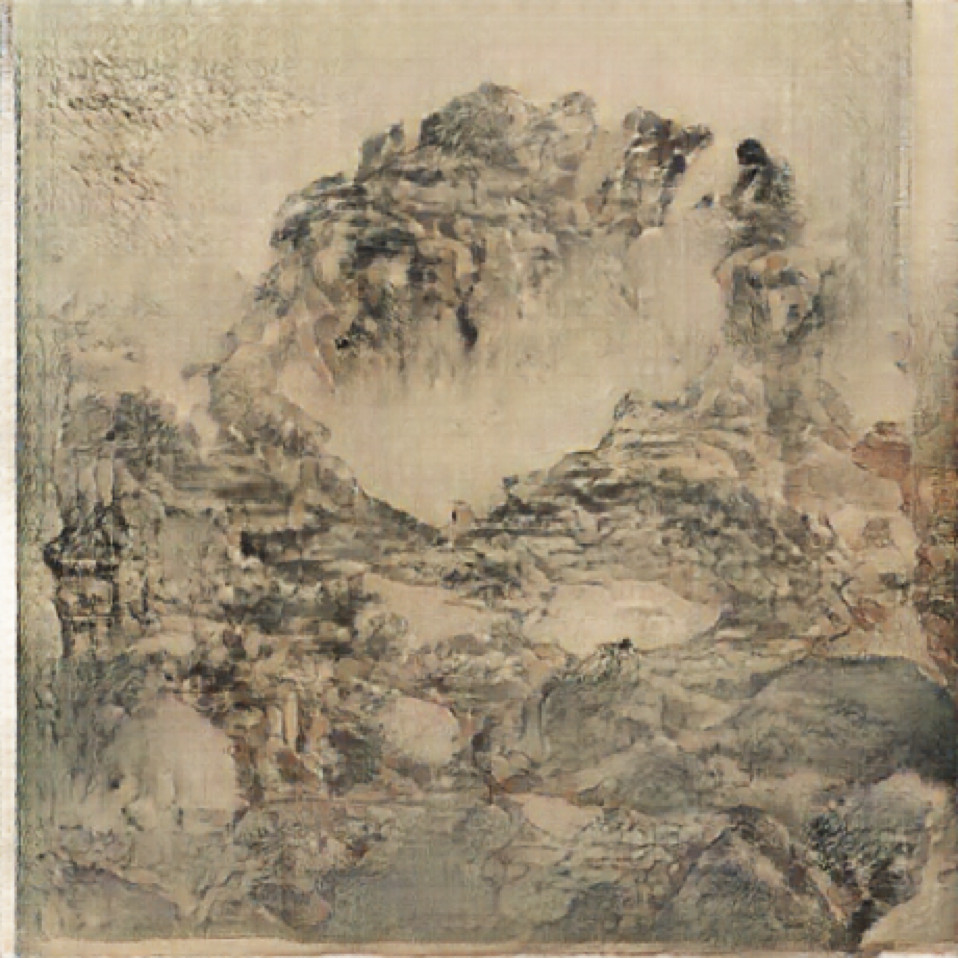} &
 \includegraphics[width=0.2\textwidth]{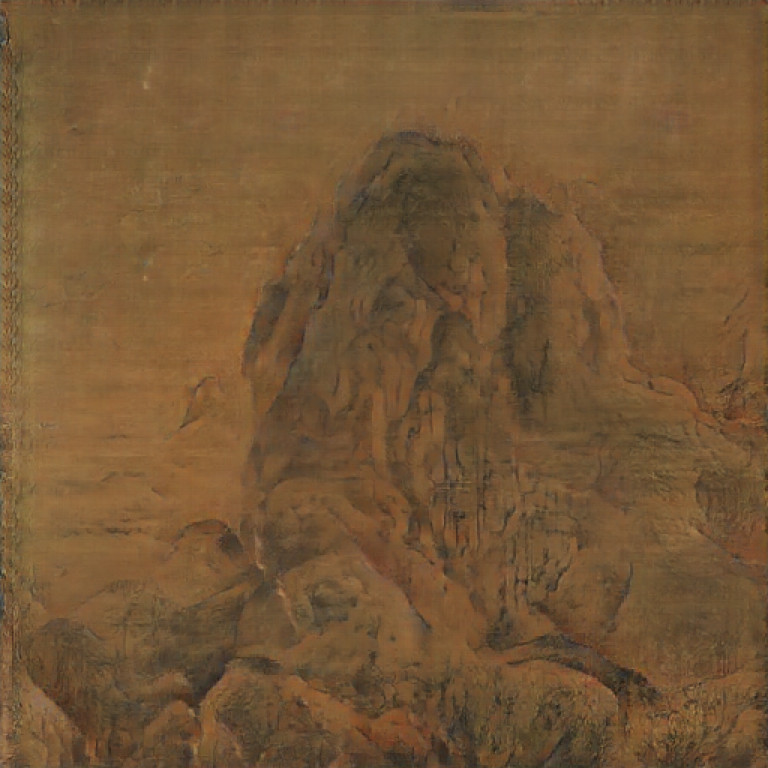}\\
 \includegraphics[width=0.2\textwidth]{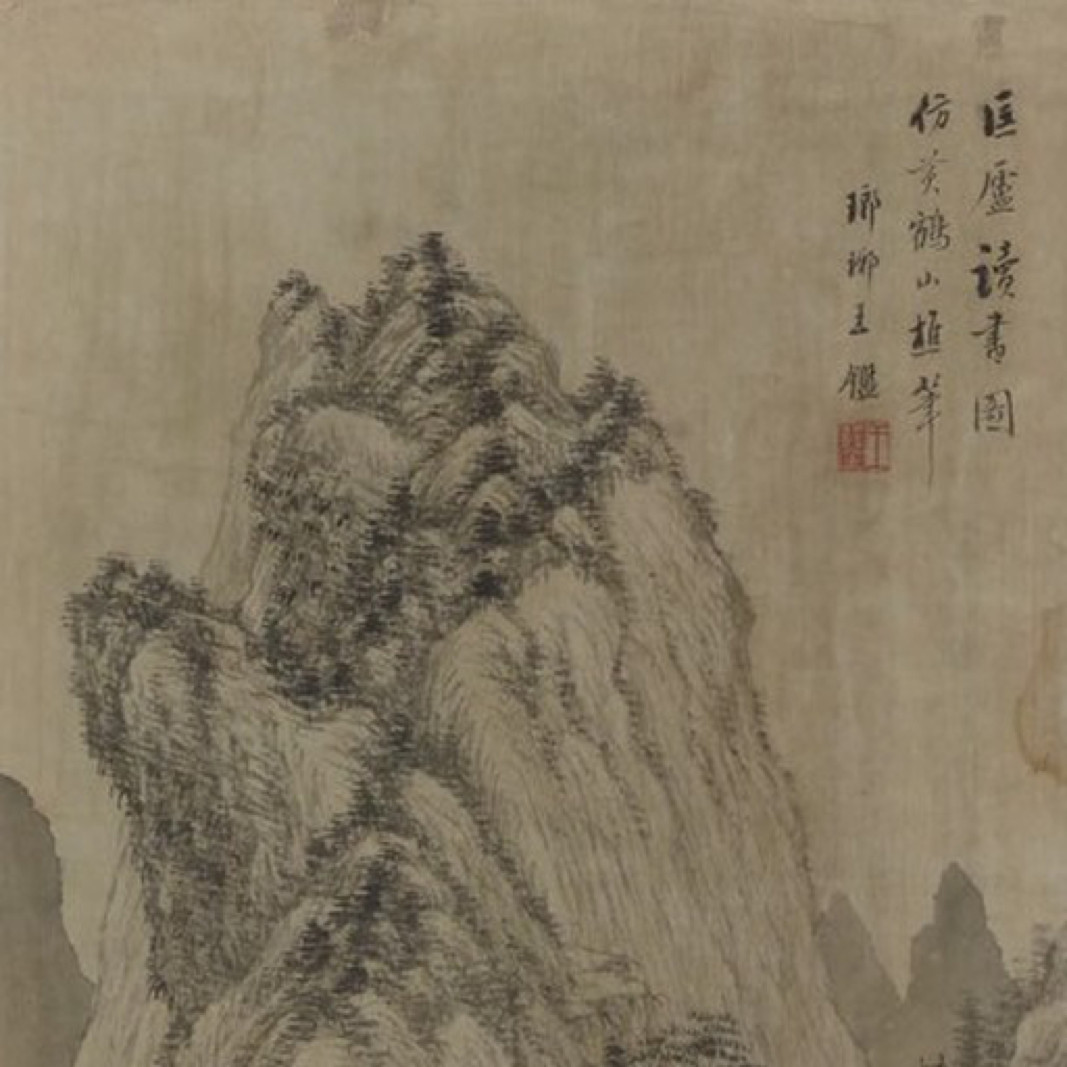} &
 \includegraphics[width=0.2\textwidth]{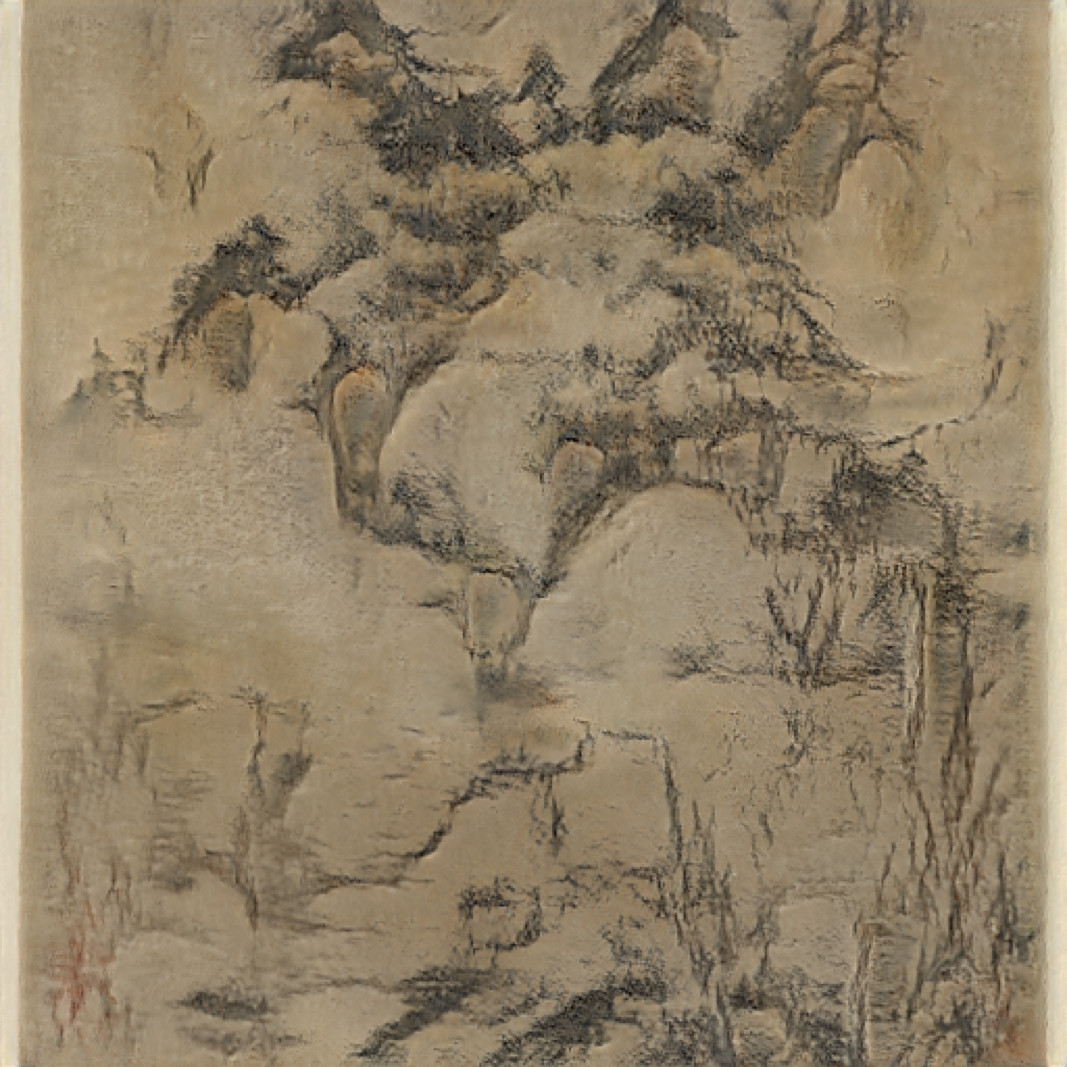} &
 \includegraphics[width=0.2\textwidth]{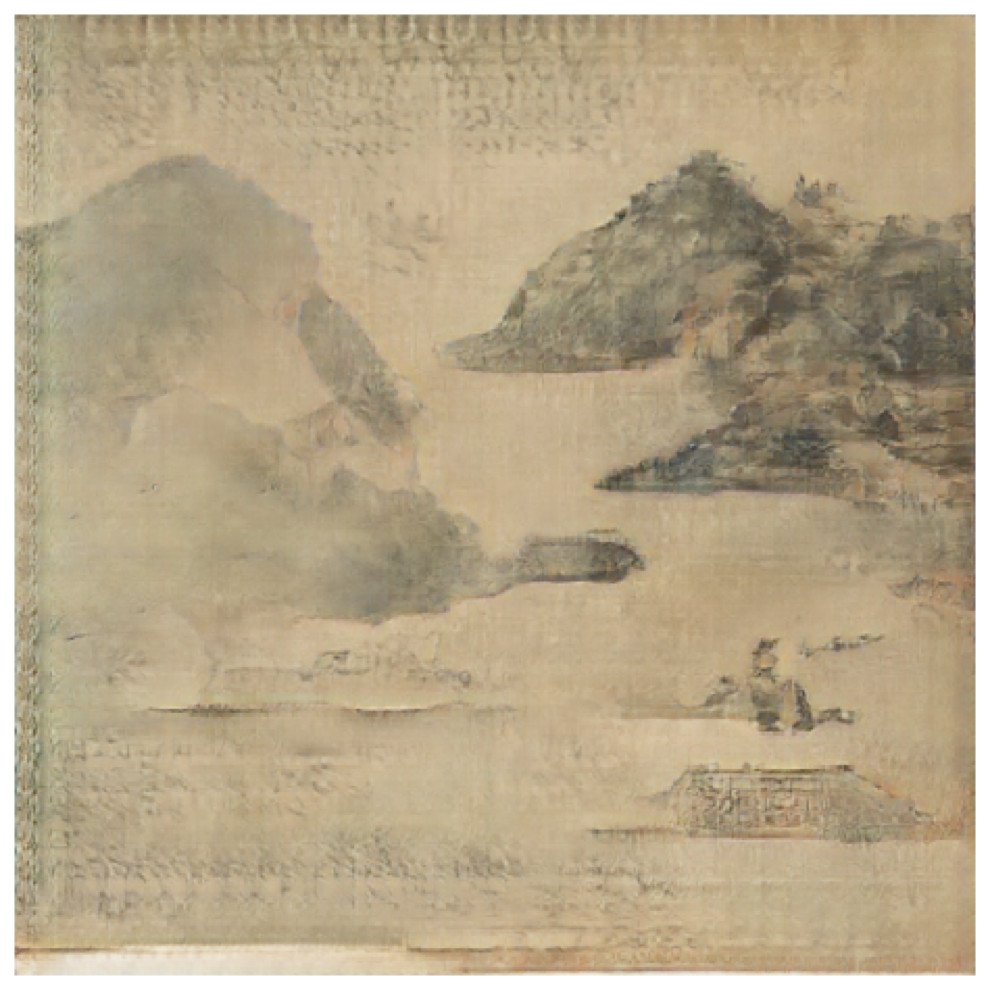} &
 \includegraphics[width=0.2\textwidth]{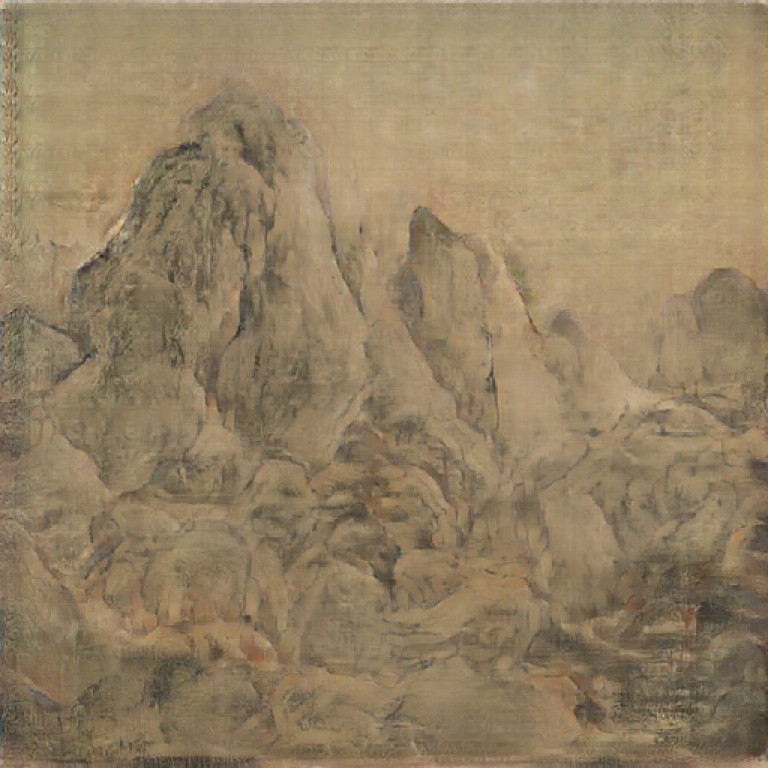}
\end{tabular}
\vspace*{2mm}
\captionof{figure}{\centering Chinese landscape paintings created by (a) human artists, (b) baseline models (top painting from RaLSGAN \cite{ralsgan}, bottom painting from StyleGAN2 \cite{stylegan2}), and two GANs, (c) and (d), within our proposed Sketch-And-Paint framework.
\label{fig:feature}}
\end{strip}

%%%%%%%%% ABSTRACT
\begin{abstract}
\vspace{10pt}
\noindent Current \ \ GAN-based \ art \ generation \ methods \ produce \ unoriginal artwork due to their dependence on conditional input. Here, we propose Sketch-And-Paint GAN (SAPGAN), the first model which generates Chinese landscape paintings from end to end, without conditional input. SAPGAN is composed of two GANs: SketchGAN for generation of edge maps, and PaintGAN for subsequent edge-to-painting translation. Our model is trained on a new dataset of traditional Chinese landscape paintings never before used for generative research. A 242-person Visual Turing Test study reveals that SAPGAN paintings are mistaken as human artwork with 55\% frequency, significantly outperforming paintings from baseline GANs. Our work lays a groundwork for truly machine-original art generation.
\end{abstract}
\vfill\null
% \columnbreak

%%%%%%%%% BODY TEXT

\section{Introduction}

Generative Adversarial Networks (GAN) have been popularly applied for artistic tasks such as turning photographs into paintings, or creating paintings in the style of modern art \cite{artgan}\cite{elgammal}. However, there are two critically underdeveloped areas in art generation research that we hope to address.

First, most GAN research focuses on Western art but overlooks East Asian art, which is rich in both historical and cultural significance. For this reason, in this paper we focus on traditional Chinese landscape paintings, which are stylistically distinctive from and just as aesthetically meaningful as Western art.

Second, popular GAN-based art generation methods such as style transfer rely too heavily on conditional inputs, e.g. photographs \cite{artgan} or pre-prepared sketches \cite{daoyu}\cite{lezhou}. There are several downsides to this. A model dependant upon conditional input is restricted in the number of images it may generate, since each of its generated images is built upon a single, human-fed input. If instead the model is not reliant on conditional input, it may generate an infinite amount of paintings seeded from latent space. Furthermore, these traditional style transfer methods can only produce derivative artworks that are stylistic copies of conditional input. In end-to-end art creation, however, the model can generate not only the style but also the content of its artworks. 

In the context of this paper, the limited research dedicated to Chinese art has not strayed from conventional style transfer methods \cite{daoyu}\cite{boli}\cite{xiaoxuan}. To our knowledge, no one has developed a GAN able to generate high-quality Chinese paintings from end to end. 
%Therefore, the purpose of this research is to achieve unsupervised generation of high-quality Chinese paintings.

Here we introduce a new GAN framework for Chinese landscape painting generation that mimics the creative process of human artists. How do painters determine their painting's composition and structure? They sketch first, then paint. Similarly, our 2-stage framework, Sketch-and-Paint GAN (SAPGAN), consists of two stages. The first-stage GAN is trained on edge maps from Chinese landscape paintings to produce original landscape "sketches," and the second-stage GAN is a conditional GAN trained on edge-painting pairs to "paint" in low-level details.

The final outputs of our model are Chinese landscape paintings which: 1) originate from latent space rather than from conditional human input, 2) are high-resolution, at 512x512 pixels, and 3) possess definitive edges and compositional qualities reflecting those of true Chinese landscape paintings. 

In summary, the contributions of our research are as follows:
\begin{itemize}
    \item We propose Sketch-and-Paint GAN, the first end-to-end framework capable of producing high-quality Chinese paintings with intelligible, edge-defined landscapes.
    \item We introduce a new dataset of 2,192 high-quality traditional Chinese landscape paintings which are exclusively curated from art museum collections. These valuable paintings are in large part untouched by generative research and are released for public usage at https://github.com/alicex2020/Chinese-Landscape-Painting-Dataset.
    \item We present experiments from a 242-person Visual Turing Test study. Results show that our model's artworks are perceived as human-created over half the time.
\end{itemize}
\vfill\null
% \columnbreak

%-------------------------------------------------------------------------
\section{Related Work}

\subsection{Generative Adversarial Networks}

The Generative Adversarial Network (GAN) consists of two models---a discriminator network $\mathcal{D}$ and a generator model $\mathcal{G}$---which are pitted against each other in a minimax two-player game \cite{goodfellow}. The discriminator's objective is to accurately predict if an input image is real or fake; the generator's objective is to fool the discriminator by producing fake images that can pass off as real. The resulting loss function is:
\begin{equation}
    \underset{\mathcal{G}}{\text{min }}\underset{\mathcal{D}}{\text{max }} \mathbb{E}_{x \sim p_{data}} [\text{log}(\mathcal{D}(x))] + \mathbb{E}_{z \sim p_{z}} [\text{log}(1 - \mathcal{D}(\mathcal{G}(z)))]
\end{equation}
where $x$ is taken from the real images denoted $p_{data}$, and $z$ is a latent vector from some probability distribution by the generator $\mathcal{G}$.

Since its inception, the GAN has been widely undertaken as a dominant research interest for generative tasks such as video frame predictions \cite{kwon}, 3D modeling \cite{hologan}, image captioning \cite{dai}, and text-to-image synthesis \cite{attngan}. Improvements to GAN distinguish between fine and coarse image representations to create high-resolution, photorealistic images \cite{huang}. Many GAN architectures are framed with an emphasis on a multi-stage, multi-generator, or multi-discriminator network distinguishing between low and high-level refinement \cite{lapgan} \cite{progressivegan} \cite{lrgan} \cite{stylegan1}. 

\subsection{Neural Style Transfer}

Style transfer refers to the mapping of a style from one image to another by preserving the content of a source image, while learning lower-level stylistic elements to match a destination style \cite{gatys}.

% The generator is fed a latent vector $z$ concatenated with a conditional input class $x$, optimizing for a loss function taking the conditional input into account (Equation \ref{eq:cganloss}) \cite{mirza}. 
% \begin{equation}
%     cGAN(\mathcal{G}, \mathcal{D}) = \mathbb{E}_{x,y}[\text{log}\mathcal{D}(x,y)] + \mathbb{E}_{x,z}[\text{log}(1 - \mathcal{D}(x, \mathcal{G}(x,z))]
%     \label{eq:cganloss}
% \end{equation}

A conditional GAN-based model called Pix2Pix performs image-to-image translation on paired data and has been popularly used for edge-to-photo image translation \cite{isola}. NVIDIA's state-of-the-art Pix2PixHD introduced photorealistic image translation operating at up to 1024x1024 pixel resolution \cite{pix2pixhd}. 

\subsubsection{Algorithmic Chinese Painting Generation}

Neural style transfer has been the basis for most published research regarding Chinese painting generation. Chinese painting generation has been attempted using sketch-to-paint translation. For instance, a CycleGAN model was trained on unpaired data to generate Chinese landscape painting from user sketches \cite{lezhou}. Other research has obtained edge maps of Chinese paintings using holistically-nested edge detection (HED), then trained a GAN-based model to create Chinese paintings from user-provided simple sketches \cite{daoyu}.

Photo-to-painting translation has also been researched for Chinese painting generation. Photo-to-Chinese ink wash painting translation has been achieved using void, brush stroke, and ink wash constraints on a GAN-based architecture \cite{chipgan}. CycleGAN has been used to map landscape painting styles onto photos of natural scenery \cite{xiaoxuan}. A mask-aware GAN was introduced to translate portrait photography into Chinese portraits in different styles such as ink-drawn and traditional realistic paintings \cite{maskaware}. However, none of these studies have created Chinese paintings without an initial conditional input like a photo or edge map.

\section{Gap in Research and Problem Formulation}

Can a computer originate art? Current methods of art generation fail to achieve true machine originality, in part due to a lack of research regarding unsupervised art generation. Past research regarding Chinese painting generation rely on image-to-image translation. Furthermore, the most popular GAN-based art tools and research are focused on stylizing existing images by using style transfer-based generative models \cite{isola}\cite{mirza}\cite{artgan}.
%In fact, the few GANs that employ end-to-end art generation characteristically generate amorphous, ill-defined images \cite{elgammal}.

Our research presents an effective model that moves away from the need for supervised input in the generative stages. Our model, SAPGAN, achieves this by disentangling content generation from style generation into two distinct networks.

To our knowledge, the most similar GAN architecture to ours is the Style and Structure Generative Adversarial Network ($\text{S}^2\text{-GAN}$) consisting of two GANs: a Structure-GAN to generate the surface normal maps of indoor scenes and Style-GAN to encode the scene's low-level details \cite{s2gan}. Similar methods have also been used in pose-estimation studies generating skeletal structures as well as mapping final appearances onto those structures \cite{villegas}\cite{yichao}.

However, there are several gaps in research that we address. First, to our knowledge, this style and structure-generating approach has never been applied to art generation. Second, we significantly optimize $\text{S}^2\text{-GAN}$'s framework with comparisons between combinations of state-of-the-art GANs such as Pix2PixHD, RaLSGAN, and StyleGAN2, which have each individually allowed for high-quality, photo-realistic image synthesis \cite{pix2pixhd}\cite{ralsgan}\cite{stylegan2}. We report a "meta" state-of-the-art model capable of generating human-quality paintings at high resolution, and outperforms current state-of-the-art models. Third, we show that generating minimal structures in the form of HED edge maps is sufficient to produce realistic images. Unlike $\text{S}^2\text{-GAN}$ (which relies on the time-intensive data collection of the XBox Kinect Sensor \cite{s2gan}) or pose estimation GANs (which are specifically tailored for pose and sequential image generation \cite{villegas}\cite{yichao}), our data processing and models are likely generalizable to any dataset encodable via HED edge detection.

%------------------------------------------------------------------------
\section{Proposed Method}

\subsection{Dataset}

We find current datasets of Chinese paintings ill-suited for our purposes for several reasons: 1) many are predominantly scraped from Google or Baidu image search engines, which often present irrelevant results; 2) none are exclusive to the traditional Chinese landscape paintings; 3) the image quality and quantity are lacking. In the interest of promoting more research in this field, we build a new dataset of high-quality traditional Chinese landscape paintings.

\noindent\textbf{Collection}. Traditional Chinese landscape paintings are collected from open-access museum galleries: the Smithsonian Freer Gallery, Metropolitan Museum of Art, Princeton University Art Museum, and Harvard University Art Museum.

\noindent\textbf{Cleaning}. We manually filter out non-landscape artworks, and hand-crop large chunks of calligraphy or silk borders out of the paintings.

\noindent\textbf{Cropping and Resizing}. Paintings are first oriented vertically and resized by width to 512 pixels while maintaining aspect ratios. A painting with a low height-to-width ratio means that the image is almost square and only a center-crop of 512x512 is needed. Paintings with a height-to-width ratio greater than 1.5 are cropped into vertical, non-overlapping 512x512 chunks. Finally, all cropped portions of reoriented paintings are rotated back to their original horizontal orientation. The final dataset counts are shown in Table \ref{table:dataset}.

\begin{table}
\begin{center}
\begin{tabular}{|l|c|}
\hline
Source & Image Count \\
\hline\hline
Smithsonian & 1,301 \\
Harvard & 101 \\
Princeton & 362 \\
Metropolitan & 428 \\
\textbf{Total} & \textbf{2,192}\\
\hline
\end{tabular}
\end{center}
\caption{Counts of images collected from four museums for our traditional Chinese landscape painting dataset}
\label{table:dataset}
\end{table}

\renewcommand{\arraystretch}{0.2}
\begin{figure}
\setlength\tabcolsep{1pt}%%
\centering
\begin{tabular}{ccc}
 \includegraphics[width=0.32\columnwidth]{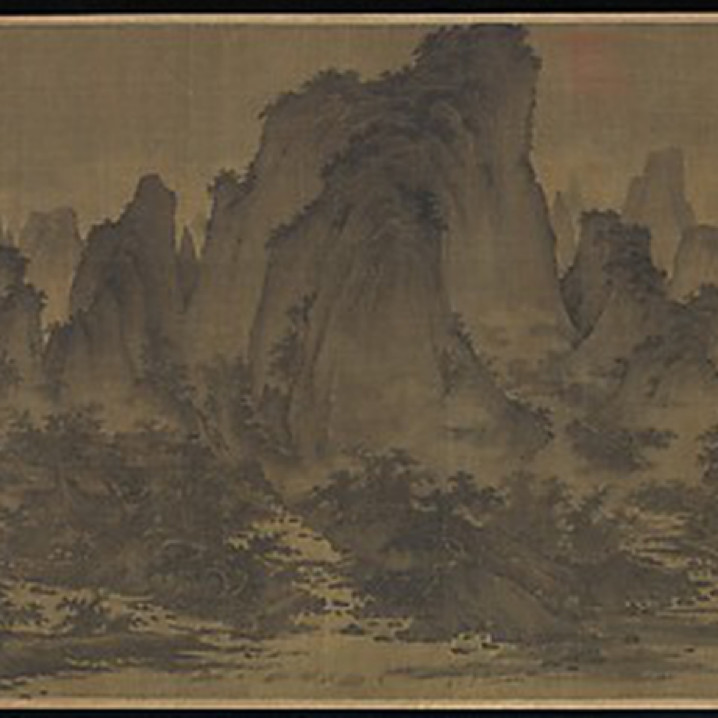} &
 \includegraphics[width=0.32\columnwidth]{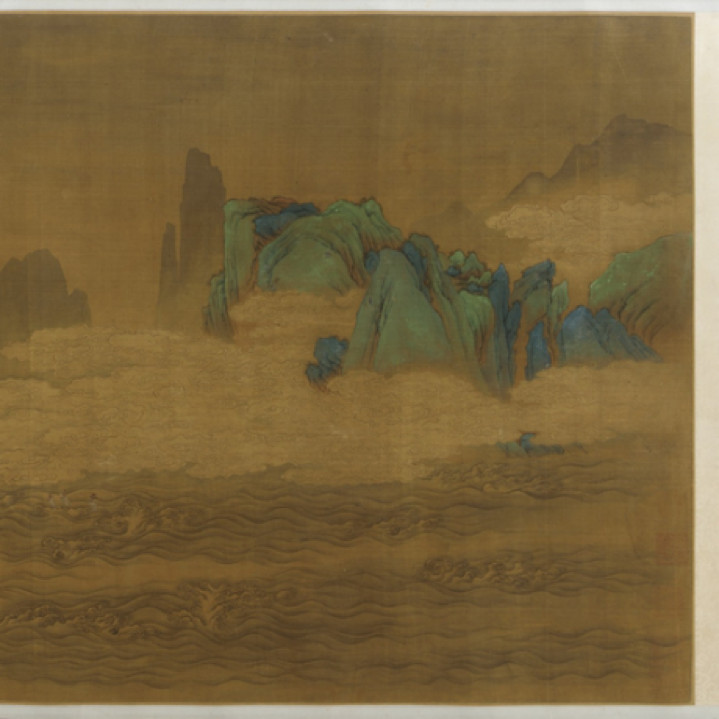} &
 \includegraphics[width=0.32\columnwidth]{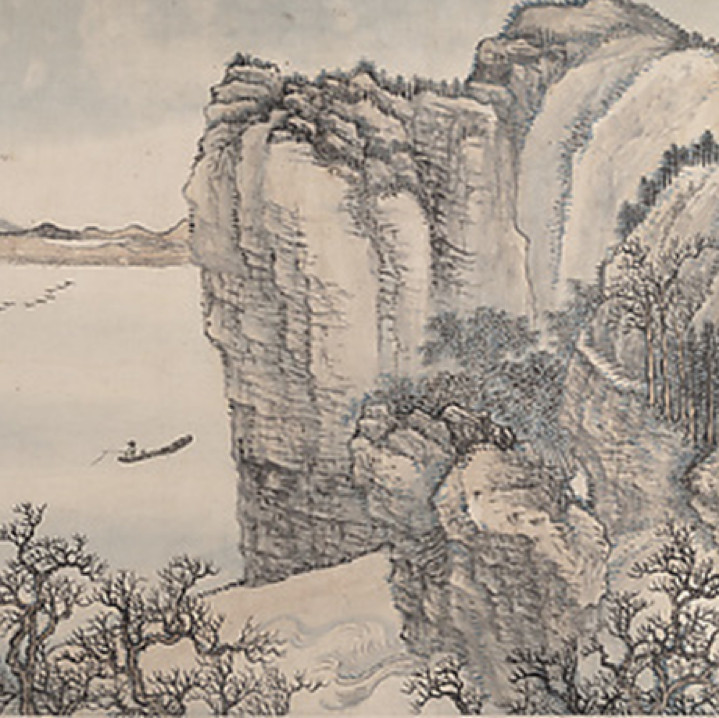}
\end{tabular}
\vspace*{1mm}
\caption{Samples from our dataset. All images are originally 512x512 pixels.}
\label{fig:samples}
\end{figure}

\noindent\textbf{Edge Maps}. HED performs edge detection using a deep learning model which consists of fully convolutional neural networks, allowing it to learn hierarchical representations of an image by aggregating edge maps of coarse-to-fine features \cite{hed}. HED is chosen over Canny edge detection due to HED's ability to clearly outline higher-level shapes while still preserving some low-level detail. We find from our experiments that Canny often misses important high-level edges as well as produces disconnected low-level edges. Thus, 512x512 HED edge maps are generated and concatenated with dataset images in preparation for training.

%------------------------------------------------------------------------
\begin{figure*}[t!]
\begin{center}
% \fbox{\rule{0pt}{2in} \rule{.9\linewidth}{0pt}}
\includegraphics[width=0.8\textwidth]{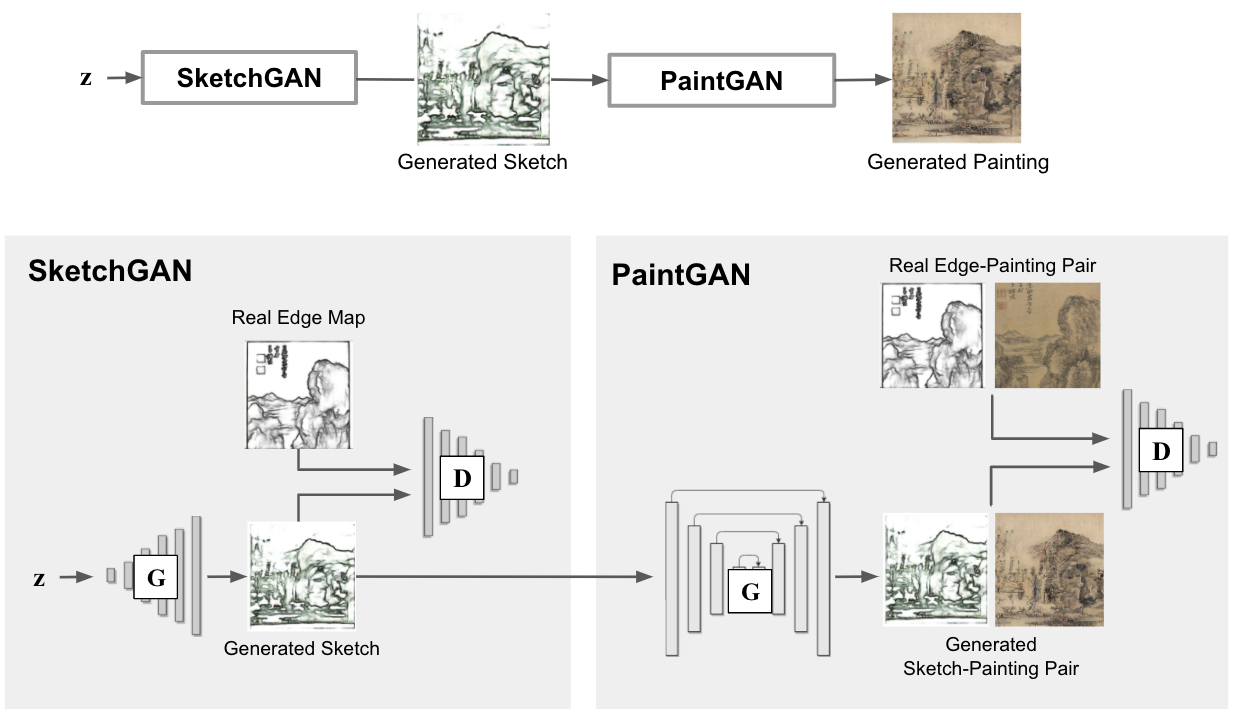}
\end{center}
   \caption{SAPGAN model framework. Top diagram shows a high-level overview of SAPGAN's generation pipeline, which starts from $z$, a latent vector. Bottom diagram details lower-level schema in which $G$ = Generator and $D$ = Discriminator. SketchGAN is trained on Chinese landscape painting edge maps. Those generated edge map are then fed to PaintGAN, which performs edge-to-painting translation to produce the final painting.}
\label{fig:model}
\end{figure*}

\subsection{Sketch-And-Paint GAN}

We propose a framework for Chinese landscape painting generation which decomposes the process into content then style generation. Our stage-I GAN, which we term "SketchGAN," generates high-resolution edge maps from a vector sampled from latent space. A stage-II GAN, "PaintGAN," is dedicated to image-to-image translation and receives the stage-I-generated sketches as input. A full model schema is diagrammed in Figure \ref{fig:model}.

Within this framework, we test different combinations of existing architectures. For SketchGAN, we train RaLSGAN and StyleGAN2 on HED edge maps. For PaintGAN, we train Pix2Pix, Pix2PixHD, and SPADE on edge-painting pairs and test these trained models on edges obtained from either RaLSGAN or StyleGAN2.

\subsubsection{Stage I: SketchGAN}

We test two models to generate HED-like edges, which serve as "sketches." SketchGAN candidates are chosen due to their ability to unconditionally synthesize high-resolution images:

\noindent\textbf{RaLSGAN.} Jolicoeur-Martineau et al. in \cite{ralsgan} introduced a relativistic GAN for high-quality image synthesis and stable training. We adopt their Relativistic Average Least-Squares GAN (RaLSGAN) and use a PACGAN discriminator (\cite{pacgan}), architecture following \cite{ralsgan}.
% Formally, if $n$ real HED edge maps $\text{\textbf{E}} = E_1, E_2, ..., E_n$ are sampled from $\mathbb{P}$, the distribution of real data, and generated edge maps $\text{\textbf{Z}} = Z_1, Z_2, ..., Z_n$ are sampled from $\mathbb{Q}$, the distribution of fake data, the SketchGAN loss function is:

% \begin{equation}
%     \begin{aligned}
%     \label{eq:edgegan}
%         &SketchGAN_\mathcal{D} = \mathbb{E}_{(E, X)\sim (\mathbb{P},\mathbb{Q})}[(D(E) - \mathbb{E}_{X}D(X) - 1)^2]\ + \\ 
%         &\ \ \ \ \ \ \ \ \ \ \ \ \ \ \ \ \ \ \ \ \ \ \ \ \  \ \mathbb{E}_{(E, Z)\sim (\mathbb{P},\mathbb{Q}}[(D(Z) - \mathbb{E}_{E}D(E) + 1)^2]\\
%         &SketchGAN_\mathcal{G} = \mathbb{E}_{(E, Z)\sim (\mathbb{P},\mathbb{Q}}[(D(Z) - \mathbb{E}_{E}D(E) - 1)^2] + \\
%         &\ \ \ \ \ \ \ \ \ \ \ \ \ \ \ \ \ \ \ \ \ \ \ \ \  \ \mathbb{E}_{(E, Z)\sim (\mathbb{P},\mathbb{Q}}[(D(E) - \mathbb{E}_{Z}D(Z) + 1)^2]
%     \end{aligned}
% \end{equation}

\noindent\textbf{StyleGAN2.} Karras et al in \cite{stylegan2} introduced StyleGAN2, a state-of-the-art model for unconditional image synthesis, generating images from latent vectors. We choose StyleGAN2 over its predecessors, StyleGAN \cite{stylegan1} and ProGAN \cite{progan}, because of its improved image quality and removal of visual artifacts arising from progressive growing. To our knowledge, StyleGAN2 has never been researched for Chinese painting generation.

\subsubsection{Stage II: PaintGAN}

PaintGAN is a conditional GAN trained with HED edges and real paintings. The following image-to-image translation models are our PaintGAN candidates.

\noindent\textbf{Pix2Pix.} Like the original implementation, we use a U-net generator and PACGAN discriminator \cite{isola}. The main change we make to the original architecture is to account for a generation of higher-resolution, 512x512 images by adding an additional downsampling and upsampling layer to the generator and discriminator. 

% Formally, if we call a batch of HED edge maps $\text{\textbf{M}} = M_1, M_2, ... , M_n$, and a batch of real paintings $\text{\textbf{X}} = X_1, X_2, ..., X_n$, where $n$ is the batch size, the L1 loss for the generator is:
% \begin{equation}
% \mathcal{L}_{L1}(G) = \mathbb{E}_{M, X}[||X - G(M)||]
% \end{equation}
% and the final loss for PaintGAN, incoporating L1 loss with a weighting parameter $\alpha$, is:
% \begin{equation}
% PaintGAN_\mathcal{G} = \underset{\mathcal{G}}{\text{min }}\underset{\mathcal{D}}{\text{max }} \mathcal{L}_{cGAN}(G, D) + \alpha (G)
% \end{equation}

\noindent\textbf{Pix2PixHD.} Pix2PixHD is a state-of-the-art conditional GAN for high-resolution, photorealistic synthesis \cite{pix2pixhd}. Pix2PixHD is composed of a coarse-to-fine generator consisting of a global and local enhancer network, and a multiscale discriminator operating at three different resolutions.

\noindent\textbf{SPADE.} SPADE is the current state-of-the-art model for image-to-image translation. Building upon Pix2PixHD, SPADE reduces the "washing-away" effect of the information encoded by the semantic map, reintroducing the input map in a spatially-adaptive layer \cite{spade}.

%------------------------------------------------------------------------
\section{Experiments}

To optimize the SAPGAN framework, we test combinations of GANs for SketchGAN and PaintGAN. In Section \ref{visual}, we assess the visual quality of individual and joint outputs from these models. In Section \ref{human_study}, we report findings from a user study.

\subsection{Training Details}

Training of the two GANs occurs in parallel: SketchGAN on edge maps generated from our dataset, and PaintGAN on edge-painting pairings. The outputs of SketchGAN are then loaded into the trained PaintGAN model.

\noindent\textbf{SketchGAN}. \textit{RaLSGAN}: The model is trained for 400 epochs. Adam optimizer is used with betas = 0.9 and 0.999, weight decay = 0, and learning rate = 0.002. \textit{StyleGAN2}: We use mirror augmentation, training from scratch for 2100 kimgs, with truncation psi of 0.5.

\noindent\textbf{PaintGAN}. \textit{Pix2Pix}: Pix2Pix is trained for 400 epochs with a batch size of 1. Adam optimizer with learning rate = 0.0002 and beta = 0.05 is use for U-net generator. \textit{Pix2PixHD:} Pix2PixHD is trained for 200 epochs with a global generator, batch size = 1, and number of generator filters = 64. \textit{SPADE:} SPADE is trained for 225 epochs with batch size of 4, load size of 512x512, and 64 filters in the generator's first convolutional layer.

\subsection{Baselines}

\noindent\textbf{DCGAN}. We find that DCGAN generates only 512x512 static noise due to vanishing gradients. No DCGAN outputs are shown for comparison, but it is an implied low baseline.

\noindent\textbf{RaLSGAN}. RaLSGAN is trained on all landscape paintings from our dataset with same configurations as listed above.

\noindent\textbf{StyleGAN2}. StyleGAN2 is trained on all landscape paintings with the same configurations as listed above.

%-------------------------------------------------------------------------
\subsection{Visual Quality Comparisons}
\label{visual}

\begin{figure}
\setlength\tabcolsep{0.5pt}%%
\centering
\begin{tabular}{cccc}
 (a) Human &
 (b) DCGAN &
 (c) RaLSGAN &
 (d) StyleGAN2\\
 \includegraphics[width=0.235\columnwidth]{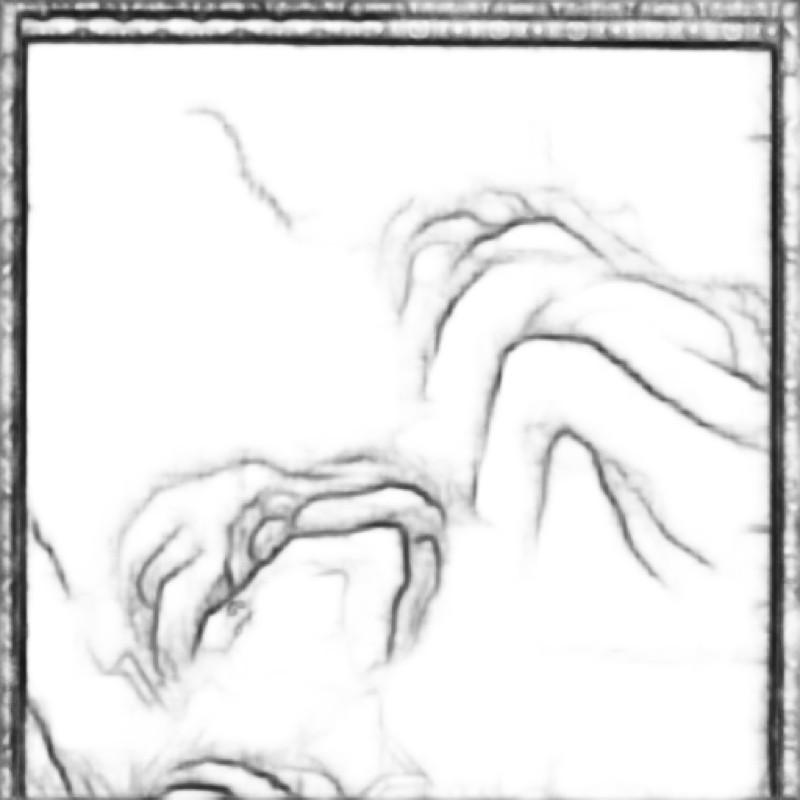} &
 \includegraphics[width=0.235\columnwidth]{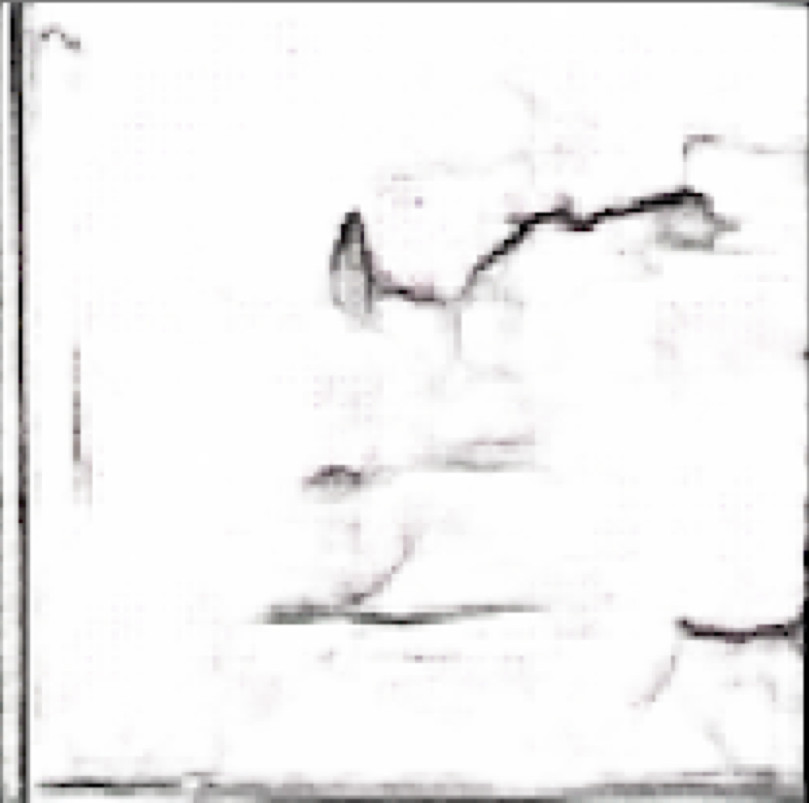} &
 \includegraphics[width=0.235\columnwidth]{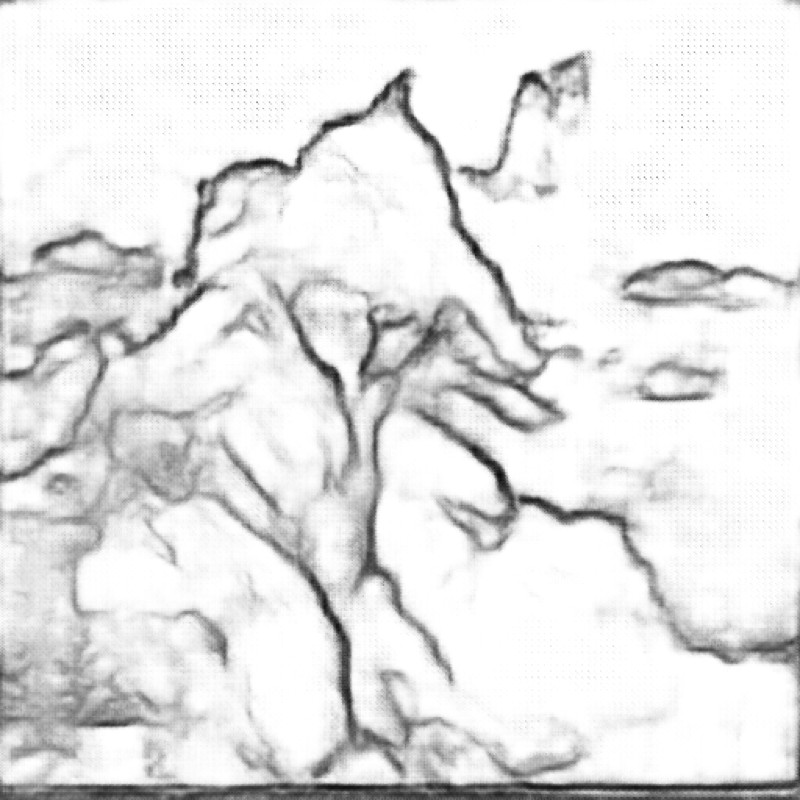} &
 \includegraphics[width=0.235\columnwidth]{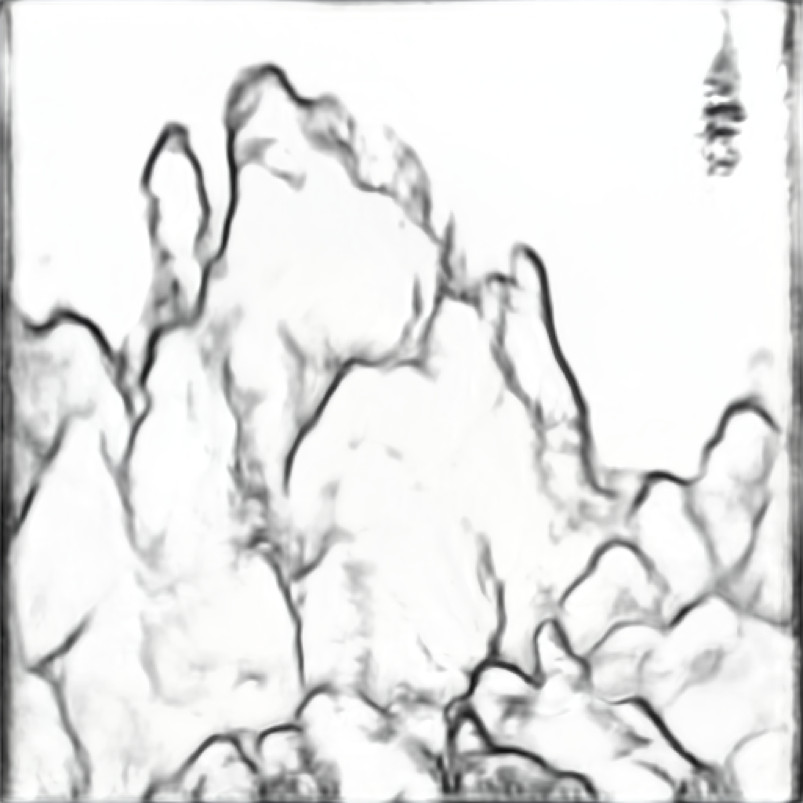}\\
  \includegraphics[width=0.235\columnwidth]{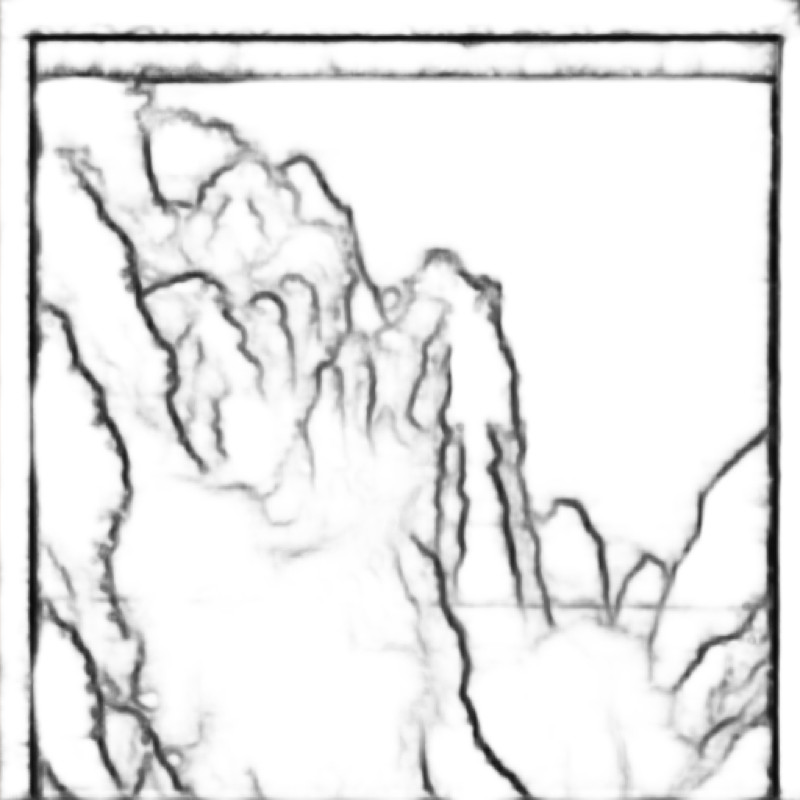} &
 \includegraphics[width=0.235\columnwidth]{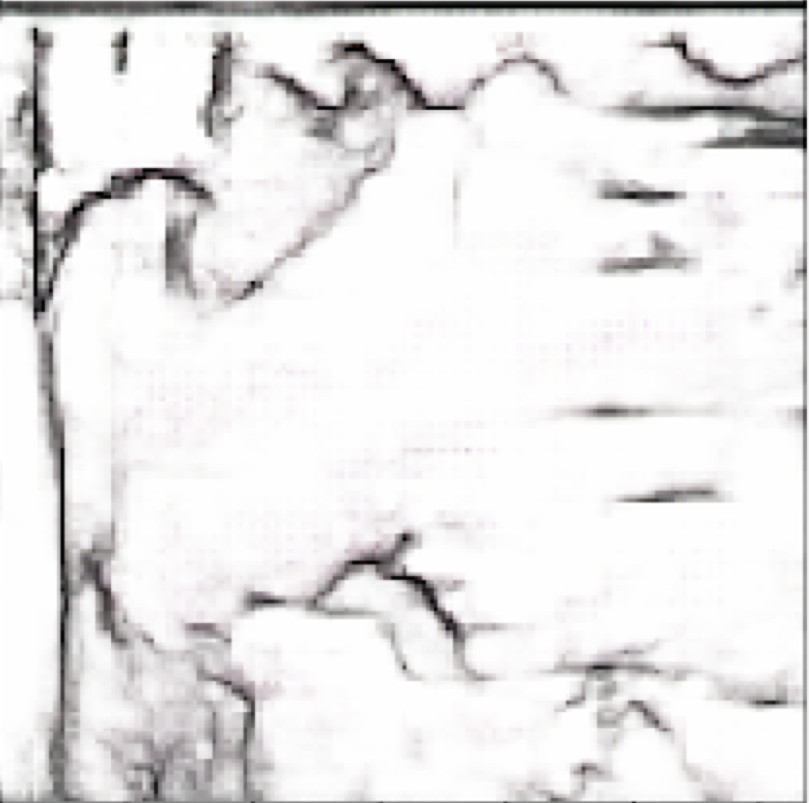} &
 \includegraphics[width=0.235\columnwidth]{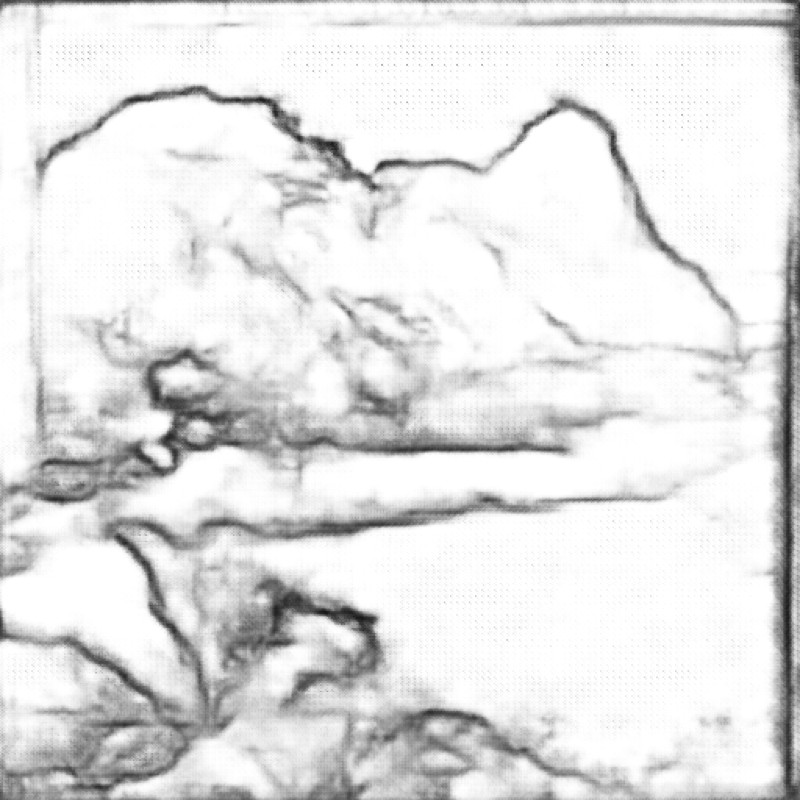} &
 \includegraphics[width=0.235\columnwidth]{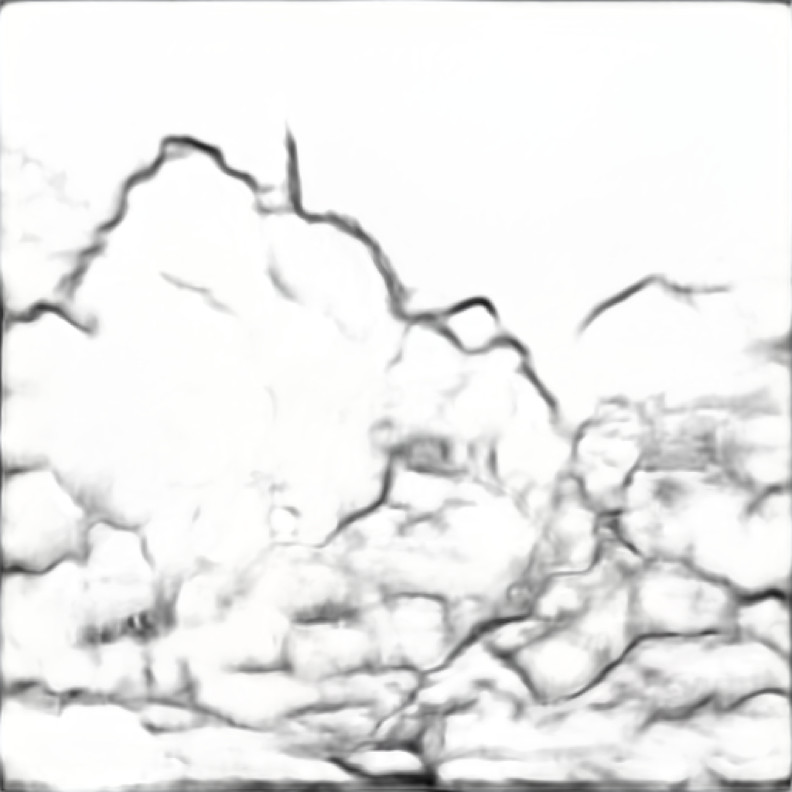}
\end{tabular}
\vspace*{0.5mm}
\caption{SketchGAN Output. Original painting HED edges (a) are compared with edges generated by SketchGAN candidate models all trained on HED edge maps: DCGAN (b), RaLSGAN (c), and StyleGAN2 (d).}
\label{fig:sketchgan}
\end{figure}

% Pix2Pix figure - not really needed
% \begin{figure}
% \setlength\tabcolsep{1pt}%%
% \centering
% \begin{tabular}{ccc}
%  (a) Input &
%  (b) GT &
%  (c) Output\\
%  \includegraphics[width=0.235\columnwidth]{images/paintgan/edge_1.png} &
%  \includegraphics[width=0.235\columnwidth]{images/paintgan/gt_1.png} &
%  \includegraphics[width=0.235\columnwidth]{images/paintgan/output_1.png} &\\
%  \includegraphics[width=0.235\columnwidth]{images/paintgan/edge_2.png} &
%  \includegraphics[width=0.235\columnwidth]{images/paintgan/gt_2.png} &
%  \includegraphics[width=0.235\columnwidth]{images/paintgan/output_2.png} &\\
% %   \includegraphics[width=0.32\columnwidth]{images/paintgan/edge_3.png} &
% %  \includegraphics[width=0.32\columnwidth]{images/paintgan/gt_3.png} &
% %  \includegraphics[width=0.32\columnwidth]{images/paintgan/output_3.png} &\\
% \end{tabular}
% \vspace*{1mm}
% \caption{Random output samples of Pix2Pix, trained on edge-painting pairs. (GT = Ground Truth)}
% \label{fig:paintgan}
% \end{figure}

\begin{figure}
\setlength\tabcolsep{1.4pt}%%
\centering\offinterlineskip
\centering
\small
\begin{tabular}{cccc}
 (a) Edge&
 (b) SPADE&
 (c) Pix2PixHD&
 (d) Pix2Pix\\
 \includegraphics[width=0.24\columnwidth]{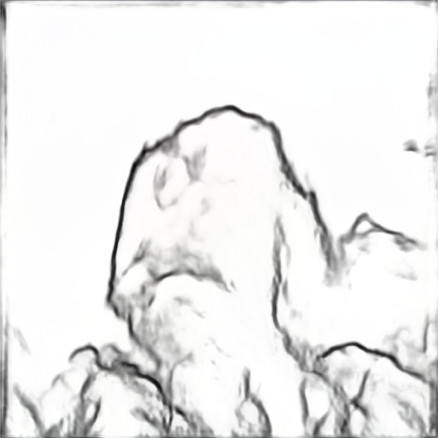} &
 \includegraphics[width=0.24\columnwidth]{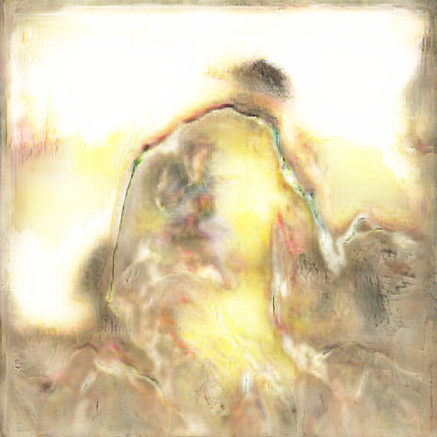} &
 \includegraphics[width=0.24\columnwidth]{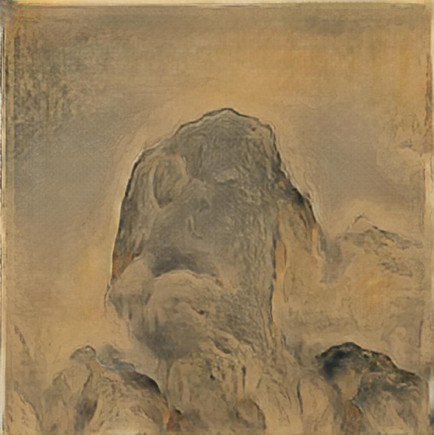} &
 \includegraphics[width=0.24\columnwidth]{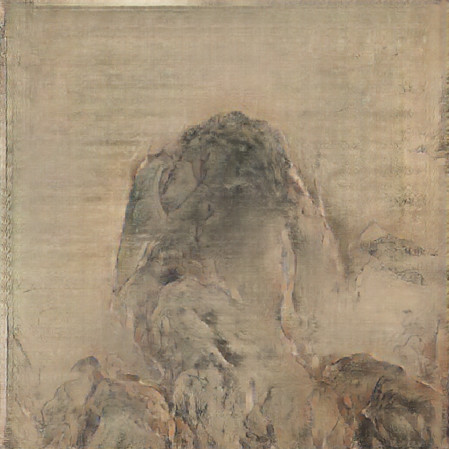} \\
 \includegraphics[width=0.24\columnwidth]{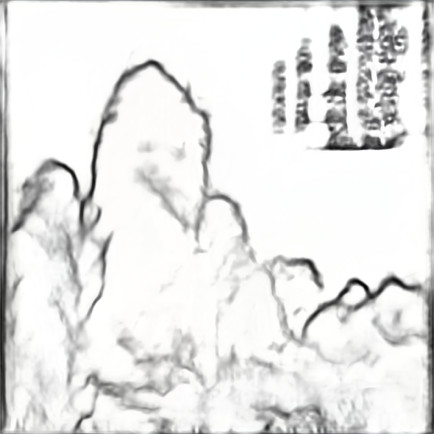} &
 \includegraphics[width=0.24\columnwidth]{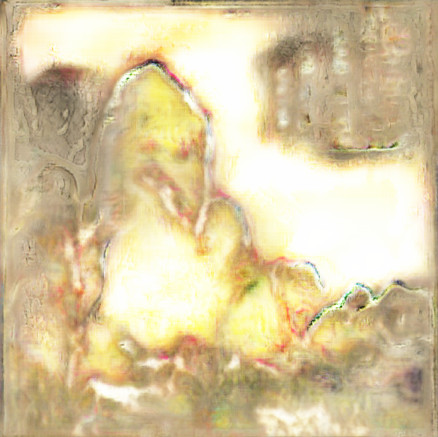} &
 \includegraphics[width=0.24\columnwidth]{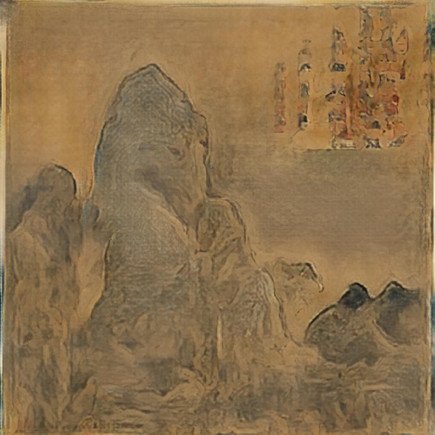} &
 \includegraphics[width=0.24\columnwidth]{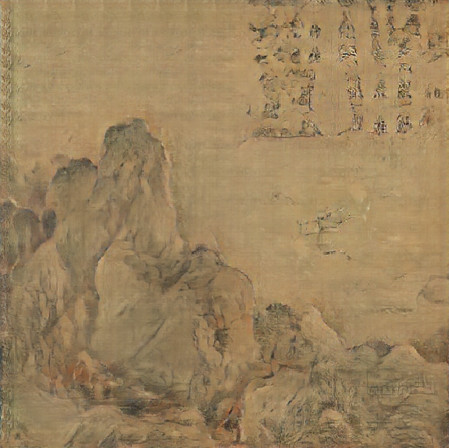}
\end{tabular}
\vspace*{1mm}
\caption{Comparisons between PaintGAN candidates fed with generated edges. (a) shows StyleGAN2-generated edges which are fed into (b) SPADE, (c) Pix2PixHD and (d) Pix2Pix.}
\label{fig:gen_edge_paint}
\end{figure}

\begin{figure*}
\setlength\tabcolsep{2.4pt}%%
\centering
\begin{tabular}{c?cc?cc}
 (a) Human&
\shortstack{(b) Baseline\\(StyleGAN2 \cite{stylegan2})}&
\shortstack{(c) Baseline\\(RaLSGAN \cite{ralsgan})}&
 \shortstack{(d) Ours\\(SAPGAN)}&
 \shortstack{(e) Ours\\(SAPGAN)}\\
 \includegraphics[width=0.19\textwidth]{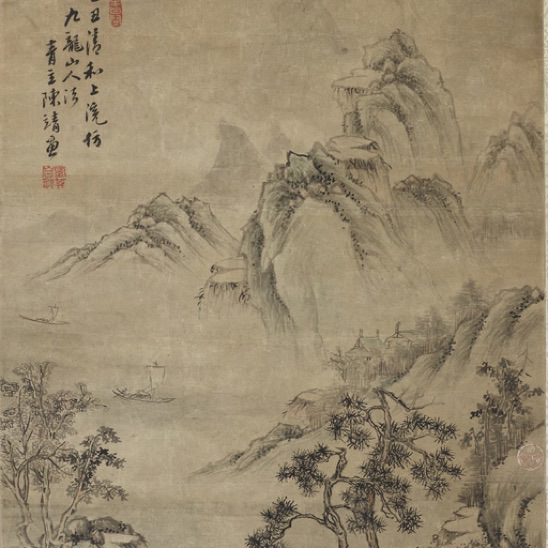} &
 \includegraphics[width=0.19\textwidth]{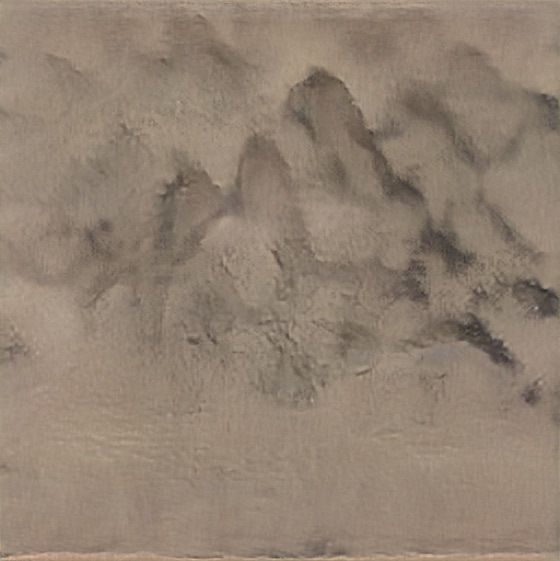} &
 \includegraphics[width=0.19\textwidth]{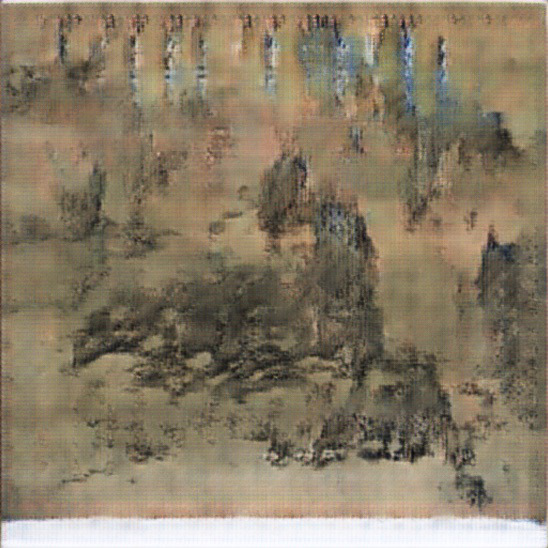} &
 \includegraphics[width=0.19\textwidth]{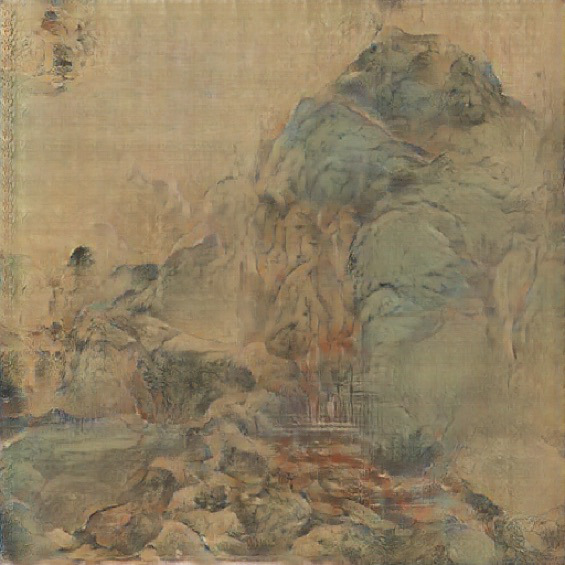} &
 \includegraphics[width=0.19\textwidth]{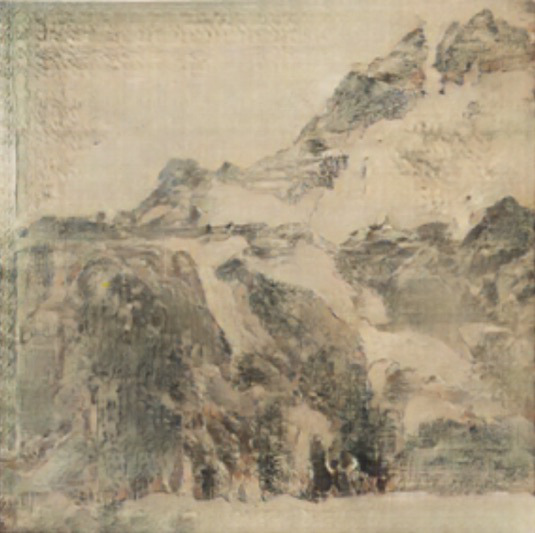}\\
  \includegraphics[width=0.19\textwidth]{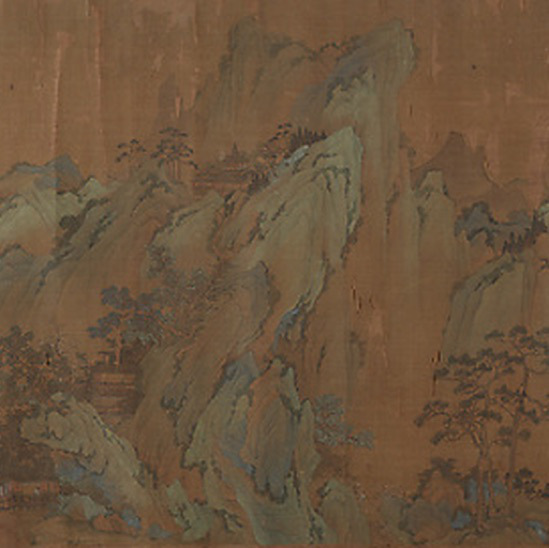} &
 \includegraphics[width=0.19\textwidth]{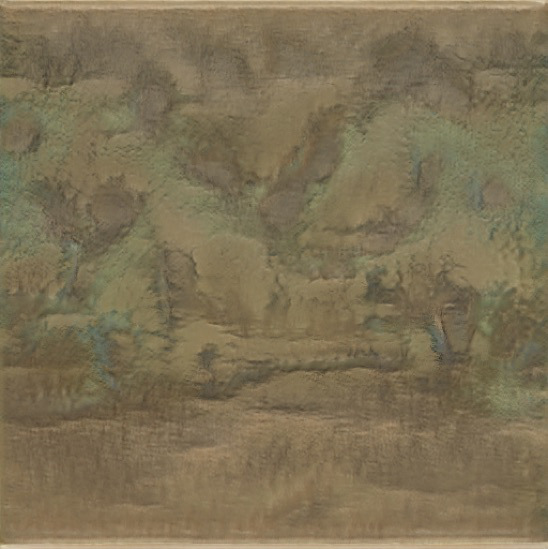} &
 \includegraphics[width=0.19\textwidth]{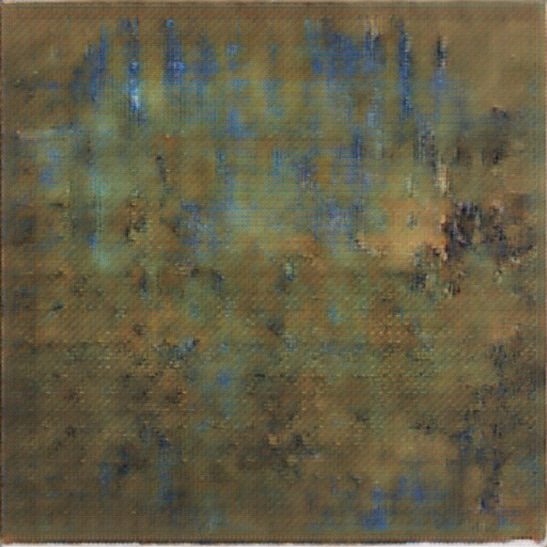} &
 \includegraphics[width=0.19\textwidth]{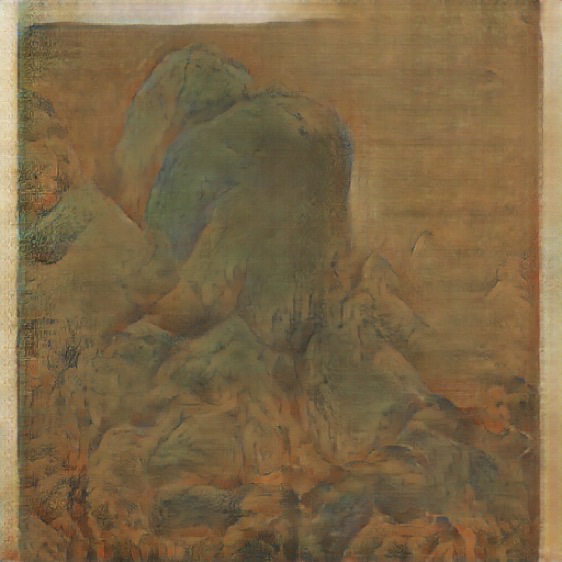} &
 \includegraphics[width=0.19\textwidth]{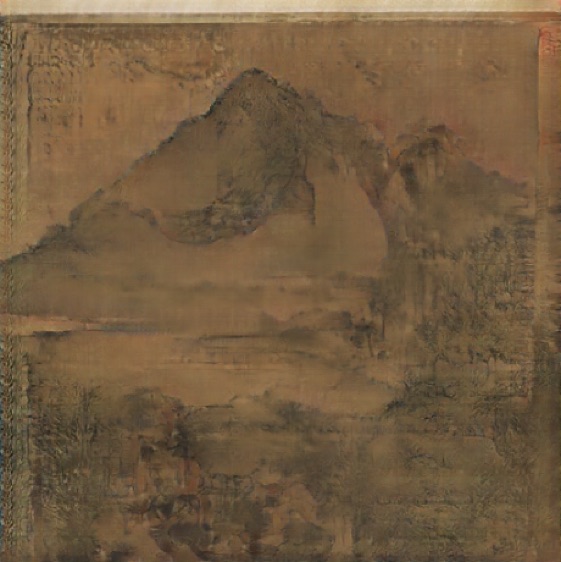}\\
  \includegraphics[width=0.19\textwidth]{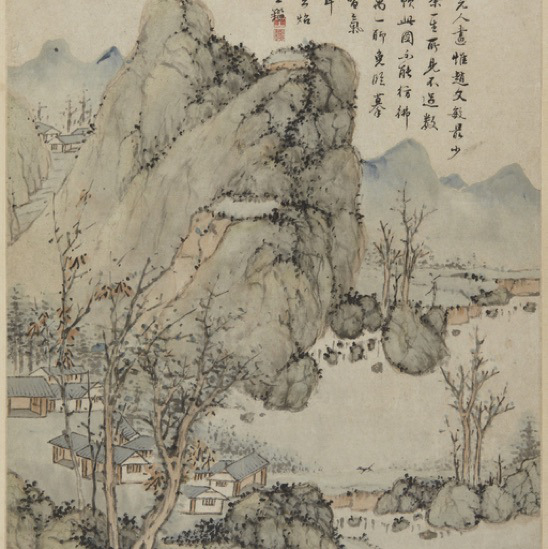} &
 \includegraphics[width=0.19\textwidth]{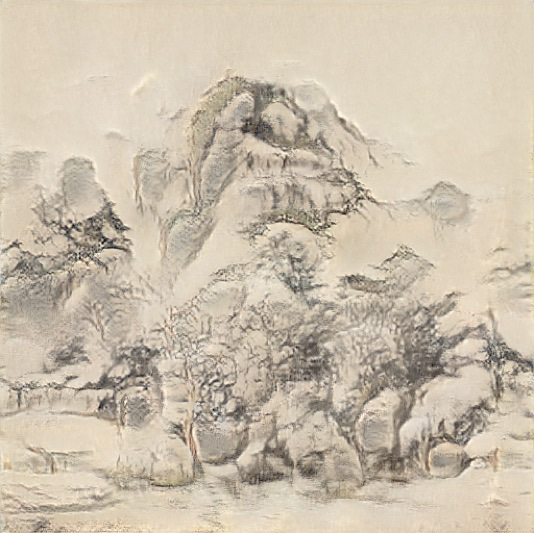} &
 \includegraphics[width=0.19\textwidth]{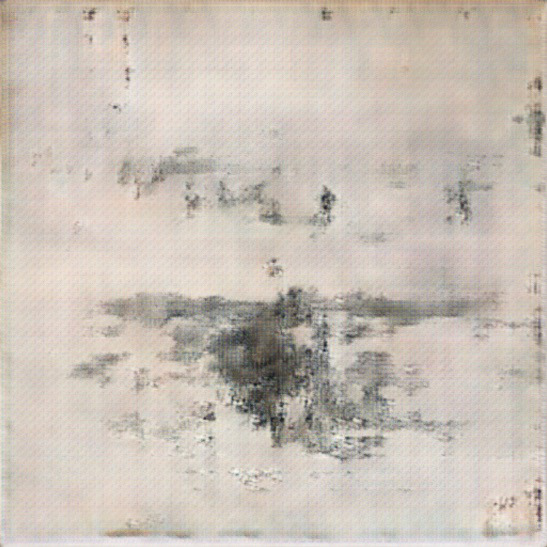} &
 \includegraphics[width=0.19\textwidth]{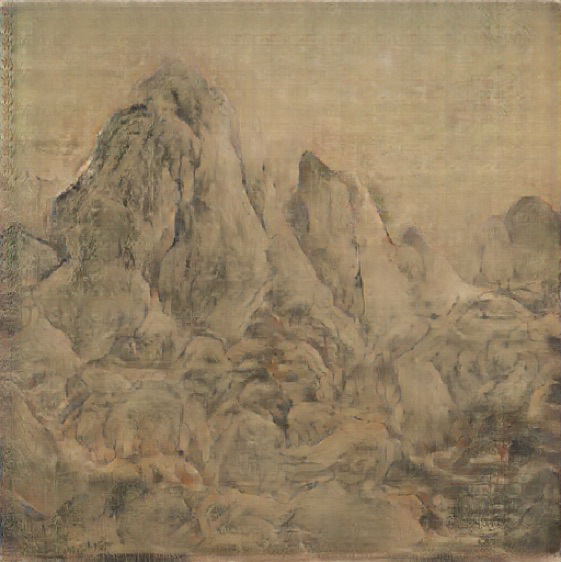} &
 \includegraphics[width=0.19\textwidth]{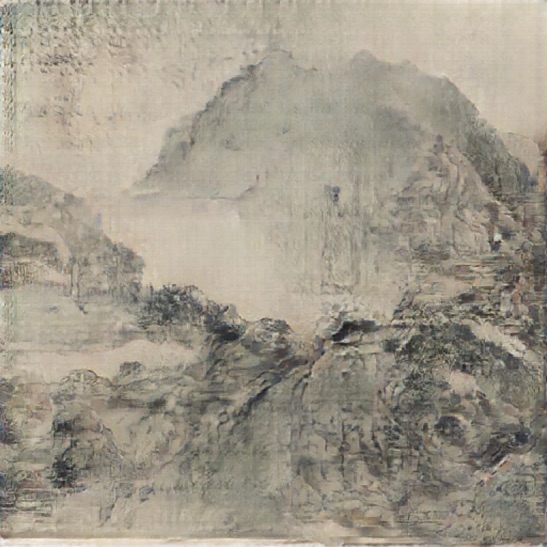}\\
\end{tabular}
\vspace{1mm}
\caption{Comparisons between Chinese landscape paintings generated by baseline models (columns b and c) versus models in our proposed Sketch-and-Paint framework (columns d and e). Specifically, the SAPGAN configurations shown are StyleGAN2+Pix2Pix (d) and RaLSGAN+Pix2Pix (e). All images are originally 512x512.}
\label{fig:all_comparisons}
\end{figure*}

\subsubsection{SketchGAN and PaintGAN Output}

We first examine the training results of SketchGAN and PaintGAN separately.

\noindent\textbf{SketchGAN.} DCGAN, RaLSGAN, and StyleGAN2 are tested for their ability to synthesize realistic edges. Figure \ref{fig:sketchgan} shows sample outputs from these models when trained on HED edge maps. DCGAN edges show little semblance of landscape definition. Meanwhile, StyleGAN and RaLSGAN outputs are clear and high-quality. Their sketches outline high-level shapes of mountains, as well as low-level details such as rocks in the terrain.

\noindent \textbf{PaintGAN.} PaintGAN candidates SPADE, Pix2PixHD, and Pix2Pix are shown in Figure \ref{fig:gen_edge_paint}. StyleGAN2-generated sketches are used as conditional input to a) SPADE, b) Pix2PixHD, and c) Pix2Pix (Figure \ref{fig:gen_edge_paint}). Noticeably, SPADE outputs' colors show evidence of over-fitting; the colors are oversaturated, yellow, and unlike those of normal landscape paintings (Figure \ref{fig:gen_edge_paint}b). Thus, we proceed further SPADE testing without SPADE. In Pix2PixHD, there are also visual artifacts, seen from the halo-like coloring around the edges of the mountains (Figure \ref{fig:gen_edge_paint}c). Pix2Pix performs the best, with fewer visual artifacts and more varied coloring. PaintGAN candidates do poorly at the granular level needed to "fill in" Chinese calligraphy, producing the blurry characters (Figure \ref{fig:gen_edge_paint}, bottom row). However, within the scope of this research, we focus on generating landscapes rather than Chinese calligraphy, which merits its own paper.

\subsubsection{Baseline Comparisons}

Both baseline models underperform in comparison to our SAPGAN models. Baseline RaLSGAN paintings show splotches of color rather than any meaningful representation of a landscape, and baseline StyleGAN2 paintings show distorted, unintelligible landscapes (Figure \ref{fig:all_comparisons}). 

Meanwhile, SAPGAN paintings are superior to baseline GAN paintings in regards to realism and artistic composition. The SAPGAN configuration, RaLSGAN edges + Pix2Pix (for brevity, the word "edges" is henceforth omitted when referencing SAPGAN models), would sometimes even separate foreground objects from background, painting distant mountains with lighter colors to establish a fading perspective (Figure \ref{fig:all_comparisons}e, bottom image). RaLSGAN+Pix2Pix also learned to paint mountainous terrains faded in mist and use negative space to represent rivers and lakes (Figure \ref{fig:all_comparisons}e, top image). The structural composition and well-defined depiction of landscapes mimic characteristics of traditional Chinese landscape paintings, adding to the paintings' realism. 

% Additional images from SAPGAN in comparison to baseline may be found in the supplementary material.

\subsubsection{Human Study: Visual Turing Tests}
\label{human_study}

We recruit 242 participants to take a Visual Turing Test. Participants are asked to judge if a painting is human or computer-created, then rate its aesthetic qualities. Among the test-takers, 29 are native Chinese speakers and the rest are native English speakers. The tests consist of 18 paintings each, split evenly between human paintings, paintings from the baseline model RaLSGAN, and paintings from SAPGAN (RaLSGAN+Pix2Pix).

For each painting, participants are asked three questions:
\begin{quote}
\textbf{Q1}: Was this painting created by a human or computer? \textit{(Human, Computer)}
\\\textbf{Q2}: How certain were you about your answer? \textit{(Scale of 1-10)}
\\\textbf{Q3}: The painting was: Aesthetically pleasing, Artfully-composed, Clear, Creative.
\textit{(Each statement has choices: Disagree, Somewhat disagree, Somewhat agree, Agree)}
\end{quote}
The Student's two-tailed t-test is used for statistical analysis, with $p < 0.05$ denoting statistical significance.

\noindent\textbf{Results.} 
Among the 242 participants, paintings from our model where mistaken as human-produced over half the time. Table \ref{table:freq} compares the frequency that SAPGAN versus baseline paintings were mistaken for human. While SAPGAN paintings passed off as human art with a 55\% frequency, the baseline RaLSGAN paintings did so only 11\% of the time ($p < 0.0001$). 

Furthermore, as Table \ref{fig:point_dist} shows, our model was rated consistently higher than baseline in all of the artistic categories: "aesthetically pleasing," "artfully-composed," "clear," and "creativity" (all comparisons $p < 0.0001$). However, in these qualitative categories, both the baseline and SAPGAN models were rated consistently lower than human artwork. The category that SAPGAN had the highest point difference from human paintings was the "Clear" category. Interestingly, though lacking in realism, baseline paintings performed best (relative to their other categories) in "Creativity"---most likely due to the abstract nature of the paintings which deviated typical landscape paintings.

We also compared results of the native Chinese- versus English-speaking participants to see if cultural exposure would allow Chinese participants to judge the paintings correctly. However, the Chinese-speaking test-takers scored 49.2\% on average, significantly lower than the English-speaking test-takers, who scored 73.5\% on average ($p < 0.0001$). Chinese speakers also mistook SAPGAN paintings for human 70\% of the time, compared with the overall 55\%. Evidently, regardless of familiarity with Chinese culture, the participants had trouble distinguishing the sources of the paintings, indicating the realism of SAPGAN-generated paintings.

\renewcommand{\arraystretch}{1}
\begin{table}
\begin{center}
\begin{tabular}{|l|c|c|}
\hline
& Average & Stddev \\
\hline\hline
Baseline & 0.11 & 0.30 \\
Ours & \textbf{0.55} (p $<$ 0.0001) & 0.17 \\
\hline
\end{tabular}
\end{center}
\vspace{1pt}
\caption{Frequency mistaken for human art by Visual Turing Test participants. Our model performs significantly better than the baseline model in fooling human evaluators.}
\label{table:freq}
\end{table}

\begin{table}[t!]
\begin{center}
\begin{tabular}{|l|c|c|c|c|}
% \begin{tabularx}{0.97\columnwidth}{XXXXX}
\hline
& Aesthetics & Composition & Clarity & Creativity \\
\hline\hline
Baseline & 1.24 & 1.25 & 1.67 & 0.90 \\
Ours & \textbf{0.35*} & \textbf{0.37*} & \textbf{0.93*} & \textbf{0.34*}\\
\hline
\end{tabular}
\end{center}
\vspace{1pt}
\caption{Average point distance of models' paintings from human paintings in qualitative categories. Points shown are on 4-point scale. Lower is better (lowest values bolded). * denotes p $<$ 0.0001}
\label{fig:point_dist}
\end{table}

\begin{figure}[t!]
\begin{center}
\includegraphics[width=\columnwidth]{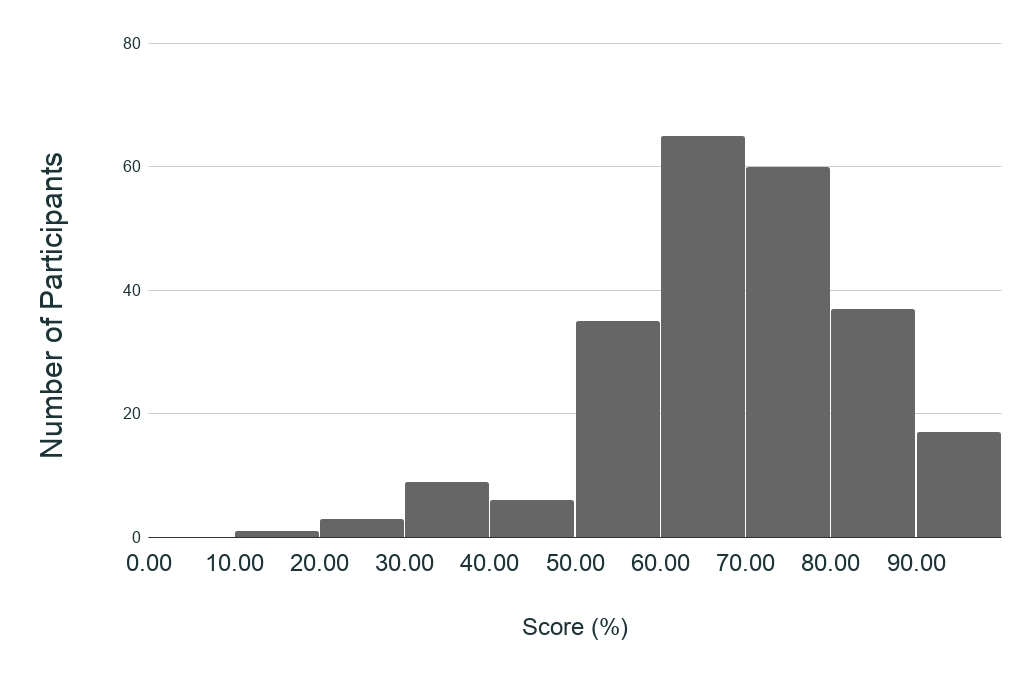}
\end{center}
\vspace*{-4pt}
   \caption{Score distribution on Visual Turing Test, asking participants to judge if an artwork was made by a human or computer (Average = 70.5\%).}
\label{fig:distribution}
\end{figure}

\begin{figure}
\setlength\tabcolsep{0.8pt}%%
\centering
\small
\begin{tabular}{ccccc}
&
 \shortstack{(a)\\Baseline}&
 \shortstack{(b)\\Baseline}&
 \shortstack{(c)\\Ours}&
 \shortstack{(d)\\Ours}\\
  \addlinespace[0.5mm]
  \multirow{1}{*}[7.5ex]{\rotatebox[origin=c]{90}{~Query}}&
 \includegraphics[width=0.185\columnwidth]{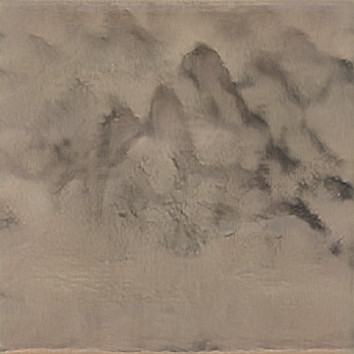} &
 \includegraphics[width=0.185\columnwidth]{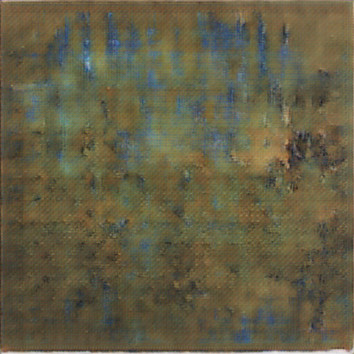} &
 \includegraphics[width=0.185\columnwidth]{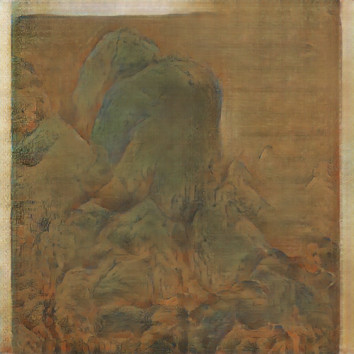} &
 \includegraphics[width=0.185\columnwidth]{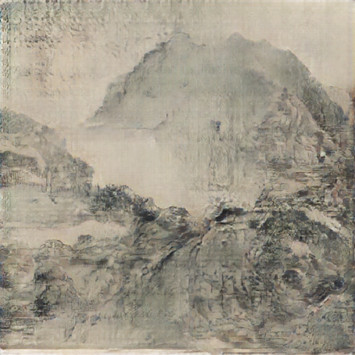}\\
   \addlinespace[0.6mm]
  \toprule
 \addlinespace[1mm]
 \multirow{3}{*}[2ex]{\rotatebox[origin=c]{90}{~Nearest Neighbors}}&
 \includegraphics[width=0.185\columnwidth]{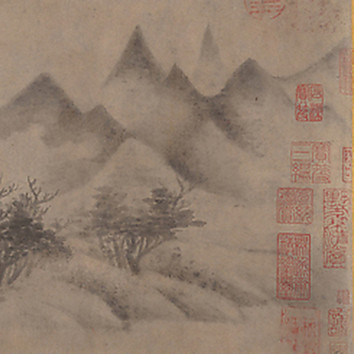} &
 \includegraphics[width=0.185\columnwidth]{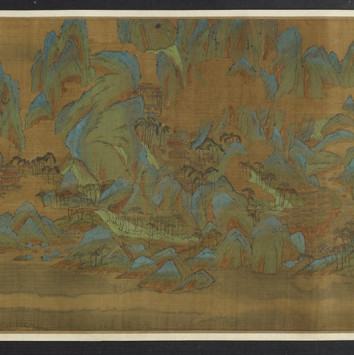} &
 \includegraphics[width=0.185\columnwidth]{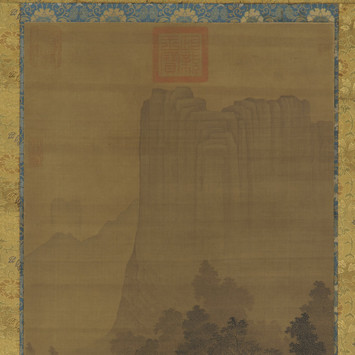} &
 \includegraphics[width=0.185\columnwidth]{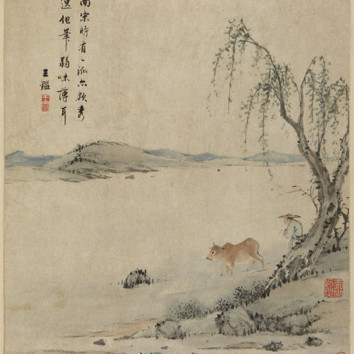}\\
 &
  \includegraphics[width=0.185\columnwidth]{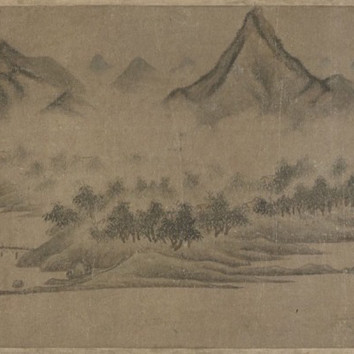} &
 \includegraphics[width=0.185\columnwidth]{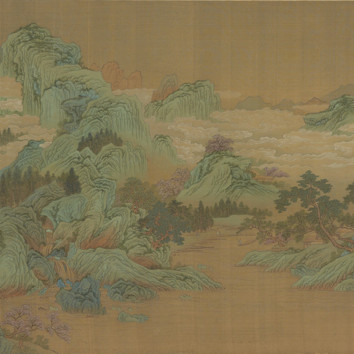} &
 \includegraphics[width=0.185\columnwidth]{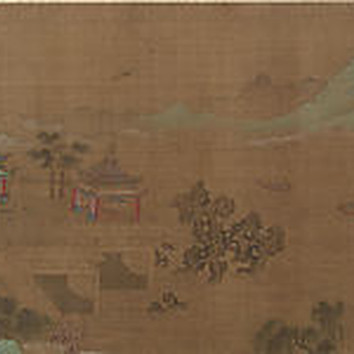} &
 \includegraphics[width=0.185\columnwidth]{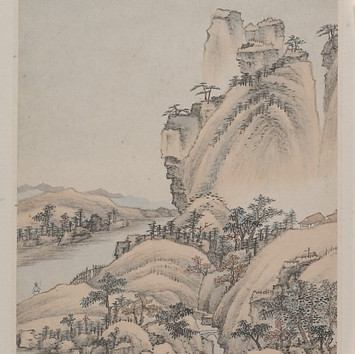}\\
  &
  \includegraphics[width=0.185\columnwidth]{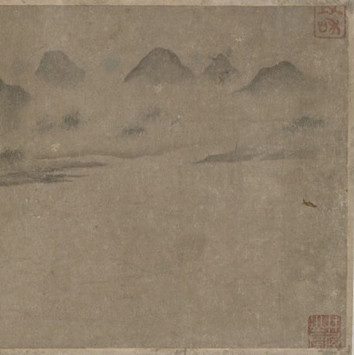} &
 \includegraphics[width=0.185\columnwidth]{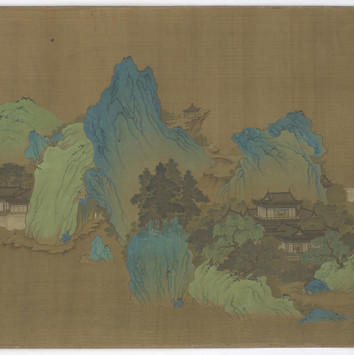} &
 \includegraphics[width=0.185\columnwidth]{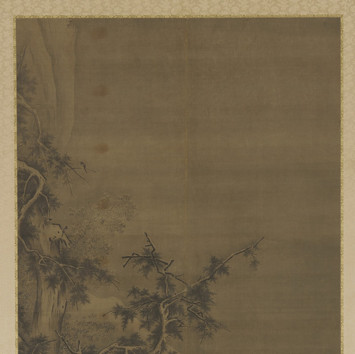} &
 \includegraphics[width=0.185\columnwidth]{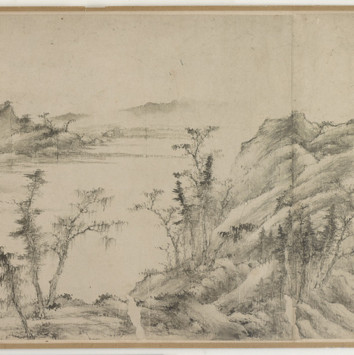}
\end{tabular}

\vspace*{1mm}
\caption{Nearest Neighbor Test. Top row shows query images outputted by (a) StyleGAN2, (b) RaLSGAN, (c) StyleGAN2+Pix2Pix (Ours), and (d) RaLSGAN+Pix2Pix (Ours). Bottom rows show the query image's three closest neighbors in the dataset.}
\label{fig:nearest_neighbors}
\end{figure}

\renewcommand{\arraystretch}{0.2}
\begin{figure}
\centering
\setlength\tabcolsep{1pt}%%
\begin{tabular}{cccccc}
 \includegraphics[width=0.158\columnwidth]{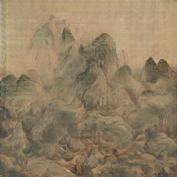} &
 \includegraphics[width=0.158\columnwidth]{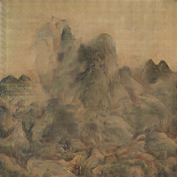} &
 \includegraphics[width=0.158\columnwidth]{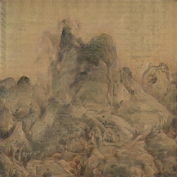} &
 \includegraphics[width=0.158\columnwidth]{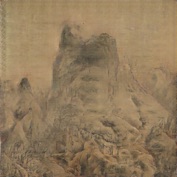} &
 \includegraphics[width=0.158\columnwidth]{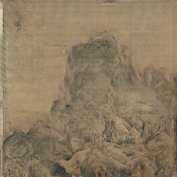} &
  \includegraphics[width=0.158\columnwidth]{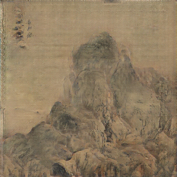}\\
   \includegraphics[width=0.158\columnwidth]{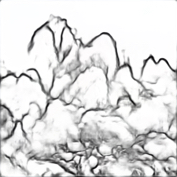} &
 \includegraphics[width=0.158\columnwidth]{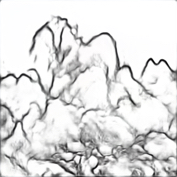} &
 \includegraphics[width=0.158\columnwidth]{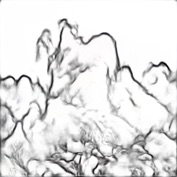} &
 \includegraphics[width=0.158\columnwidth]{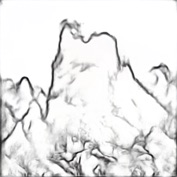} &
 \includegraphics[width=0.158\columnwidth]{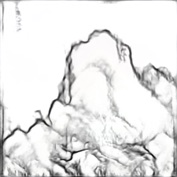} &
  \includegraphics[width=0.158\columnwidth]{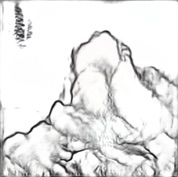}\\
  \includegraphics[width=0.158\columnwidth]{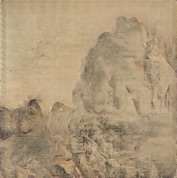} &
 \includegraphics[width=0.158\columnwidth]{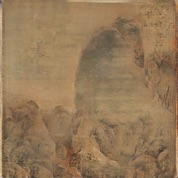} &
 \includegraphics[width=0.158\columnwidth]{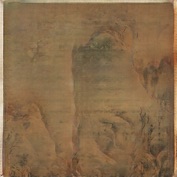} &
 \includegraphics[width=0.158\columnwidth]{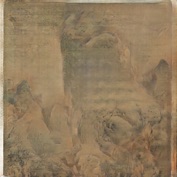} &
 \includegraphics[width=0.158\columnwidth]{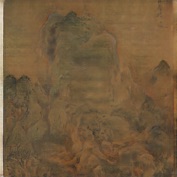} &
  \includegraphics[width=0.158\columnwidth]{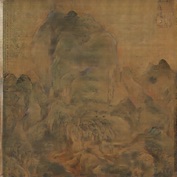}\\
  \includegraphics[width=0.158\columnwidth]{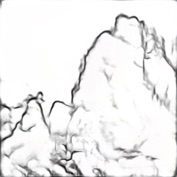} &
 \includegraphics[width=0.158\columnwidth]{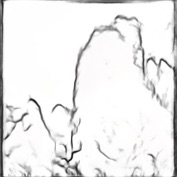} &
 \includegraphics[width=0.158\columnwidth]{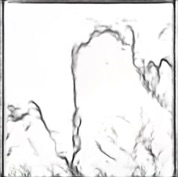} &
 \includegraphics[width=0.158\columnwidth]{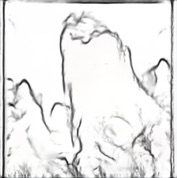} &
 \includegraphics[width=0.158\columnwidth]{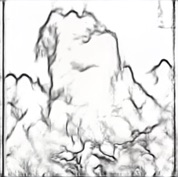} &
  \includegraphics[width=0.158\columnwidth]{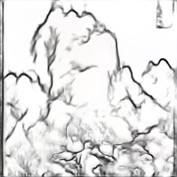}
\end{tabular}
\vspace*{0.5mm}
\caption{Latent walks from SAPGAN (StyleGAN2+Pix2Pix). StyleGAN2 sketches are shown in rows 2 and 4; their final paintings from Pix2Pix are shown in rows 1 and 3.}
\label{fig:latent}
\end{figure}

\subsection{Nearest Neighbor Test}

The Nearest Neighbor Test is used to judge a model's ability to deviate from its training dataset. To find a query's closest neighbors, we compute pixel-wise $L2$ distances from the query image to each image in our dataset. Results show that baselines, especially StyleGAN2, produce output that is visually similar to training data. Meanwhile, paintings produced by our models creatively stray from the original paintings (Figure \ref{fig:nearest_neighbors}). Thus, unlike baseline models, SAPGAN does not memorize its training set and is robust to over-fitting, even on a small dataset.

\subsection{Latent Interpolations}

Latent walks are shown to judge the quality of interpolations by SAPGAN (Figure \ref{fig:latent}). With SketchGAN (StyleGAN2), we first generate six frames of sketch interpolations from two random seeds, then feed them into PaintGAN (Pix2Pix) to generate interpolated paintings. Results show that our model can generate paintings with intelligible landscape structures at every step, most likely due to the high quality of StyleGAN2's latent space interpolations as reported in \cite{stylegan2}.

\section{Future Work}

Future work may substitute different GANs for SketchGAN and PaintGAN, allowing for more functionality such as multimodal generation of different painting styles \cite{bicyclegan}. Combinations GANs that are capable of adding brushstrokes or calligraphy onto the generated paintings may also increase appearances of authenticity \cite{lyu}. 

Importantly, apart from being trained on a Chinese landscape painting dataset, our proposed model is not specifically tailored to Chinese paintings and may be generalized to other artistic styles which also emphasize edge definition. Future work may test this claim.

\section{Conclusion}

We propose the first model that creates high-quality Chinese landscape paintings from scratch. Our proposed framework, Sketch-And-Paint GAN (SAPGAN), splits the generation process into sketch generation to create high-level structures, and paint generation via image-to-image translation. Visual quality assessments find that paintings from the RaLSGAN+Pix2Pix and StyleGAN2+Pix2Pix configurations for SAPGAN are more edge-defined and realistic in comparison to baseline paintings, which fail to evoke intelligible structures. SAPGAN is trained on a new dataset of exclusively museum-curated, high-quality traditional Chinese landscape paintings. Among 242 human evaluators, SAPGAN paintings are mistaken for human art over half of the time (55\% frequency), significantly higher than that of paintings from baseline models. SAPGAN is also robust to over-fitting compared with baseline GANs, suggesting that it can creatively deviate from its training images. Our work supports the possibility that a machine may originate artworks. 

\section{Acknowledgments}

We thank Professor Brian Kernighan, the author's senior thesis advisor, for his guidance and mentorship throughout the course of this research. We also thank Princeton Research Computing for computing resources, and the Princeton University Computer Science Department for funding.

{\small
\bibliographystyle{ieee_fullname}
\bibliography{egbib}
}

\end{document}